\documentclass[runningheads]{llncs}

\usepackage{epsfig}
\usepackage{graphicx}
\usepackage{amsmath}
\usepackage{amssymb}
\usepackage{tikz}
\usepackage{comment}
\usepackage{cite}
\usepackage{xspace}

\usepackage{booktabs}
\usepackage{amsmath, bm, subfigure, epstopdf, url, pifont, overpic, cases}
\usepackage{latexsym, amssymb, bbding, multirow, makecell,  diagbox, enumitem, caption}
\usepackage{array, enumitem, soul, algorithm, algpseudocode}
\usepackage{xcolor, color, colortbl}
\usepackage{arydshln}
\usepackage{multirow, multicol}
\usepackage{adjustbox}
\usepackage{url}
\usepackage[pagebackref,breaklinks,colorlinks]{hyperref}

\newcommand{\algname}{Swin2SR}
\newcommand{\ours}{Swin2SR }
\newcommand{\R}[1]{\textcolor[rgb]{1.00,0.00,0.00}{#1}}
\newcommand{\B}[1]{\textcolor[rgb]{0.00,0.00,1.00}{#1}}
\definecolor{Gray}{gray}{0.93}

\makeatletter
\DeclareRobustCommand\onedot{\futurelet\@let@token\@onedot}
\def\@onedot{\ifx\@let@token.\else.\null\fi\xspace}

\def\ie{\emph{i.e}\onedot}

\def\et{\emph{et al}\onedot}
\makeatother

\usepackage[accsupp]{axessibility}  

\begin{document}
\pagestyle{headings}
\mainmatter
\def\ECCVSubNumber{65}  

\title{Swin2SR: SwinV2 Transformer for Compressed Image Super-Resolution and Restoration}

\titlerunning{Swin2SR}
%
\author{Marcos~V.~Conde\inst{1} \and Ui-Jin Choi\inst{2} \and Maxime Burchi\inst{1} \and Radu Timofte\inst{1}}
\authorrunning{Conde and Choi et al.}
%
\institute{
Computer Vision Lab, CAIDAS, University of Würzburg, Germany
\email{\{marcos.conde-osorio,radu.timofte\}@uni-wuerzburg.de}\\
\and
MegaStudyEdu, South Korea
}

\maketitle

\begin{abstract}
Compression plays an important role on the efficient transmission and storage of images and
videos through band-limited systems such as streaming services, virtual reality or videogames.
However, compression unavoidably leads to artifacts and the loss of the original information, which may severely degrade the visual quality. For these reasons, quality enhancement of compressed images has become a popular research topic.
While most state-of-the-art image restoration methods are based on convolutional neural networks, other transformers-based methods such as SwinIR, show impressive performance on these tasks. 

In this paper, we explore the novel Swin Transformer V2, to improve SwinIR for image super-resolution, and in particular, the compressed input scenario. Using this method we can tackle the major issues in training transformer vision models, such as training instability, resolution gaps between pre-training and fine-tuning, and hunger on data. 
We conduct experiments on three representative tasks: JPEG compression artifacts removal, image super-resolution (classical and lightweight), and compressed image super-resolution.
Experimental results demonstrate that our method, Swin2SR, can improve the training convergence and performance of SwinIR, and is a top-5 solution at the ``AIM 2022 Challenge on Super-Resolution of Compressed Image and Video". 

Our code can be found at \url{https://github.com/mv-lab/swin2sr}.

\end{abstract}

\section{Introduction}
\label{sec:intro}

Compression plays an important role on the efficient transmission and storage of images and videos through band-limited systems such as streaming services, virtual reality, cloud storage for images, videoconferences or videogames.
However, compression leads to artifacts and the loss of the original information, which may severely degrade the visual quality of the image. For these reasons, quality enhancement and restoration of compressed images has become a popular research topic.
Image restoration techniques, such as image super-resolution (SR) and JPEG compression artifact reduction, aim to reconstruct the high-quality clean image from its low-quality degraded (or compressed) counterpart. 
During the past decade, several revolutionary works were proposed for single image super-resolution, most of them are CNN-based methods~\cite{dong2014srcnn, kim2016vdsr, zhang2017DnCNN, zhang2017IRCNN, wang2018esrgan, zhang2018ffdnet, zhang2018rcan, fritsche2019frequency, zhang2018srmd, li2019SRFBN, kai2021bsrgan, zhang2021DPIR}. 
We can also find plenty of proposed methods for the reduction of JPEG artifacts~\cite{tai2017memnet,ehrlich2020quantization,jiang2021towards}.
Recently, the blind super-resolution~\cite{gu2019sftmdikc,yamac2021kernelnet,kai2021bsrgan} methods have been proposed. They are able to use one model to jointly handle the tasks of super-resolution, deblurring, JPEG artifacts reduction, etc.
Although the performance of these deep learning methods significantly improved compared with traditional methods~\cite{timofte2014a}, they generally suffer from two basic problems that arise from the basic convolution layer receptive field: (i) the interactions between images and kernels are content-independent, therefore, using the same kernel to restore different image regions may not be the best. (ii) Under the principle of locality, convolution is not effective for long-range dependency modelling~\cite{liang2021swinir}.

As an alternative to CNNs, Transformer~\cite{vaswani2017transformer} designs a self-attention mechanism to capture global interactions between contexts and has shown promising performance in several vision problems~\cite{carion2020DETR, touvron2020DeiT, dosovitskiy2020ViT, liu2021swin}. Recently, Swin Transformer~\cite{liu2021swin} has shown great promise as it leverages the advantages of both CNN and Transformers (\ie{ CNN to process image with large size due to the local attention mechanism, and transformer to model long-range dependency with the shifted window scheme}).
Compared with classical CNN-based image restoration models, Transformer-based methods have several benefits: (i) content-based interactions between image content and attention weights, which can be interpreted as spatially varying convolution~\cite{vaswani2021SAhaloing}. (ii) long-range dependency modelling are enabled by the shifted window mechanism. (iii) in some cases, better performance with less parameters. 
In this context, Liang \et{} SwinIR~\cite{liang2021swinir}, based on Swin Transformer~\cite{liu2021swin}, represents the state-of-the-art of transformer-based models for image restoration.
 
\vspace{-2mm}
 
\paragraph{\textbf{AIM 2022 challenge on Super-Resolution of Compressed Image and Video}}

This challenge is a step forward for establishing a new benchmark for the super-resolution of JPEG images and videos. The methods proposed in this challenge also have the potential to solve various super-resolution tasks. The challenge utilizes the famous DIV2K~\cite{agustsson2017ntire} dataset for evaluating methods.
Other related challenges such as ``NTIRE 2022 challenge on super-resolution and quality enhancement of compressed video"~\cite{yang2022ntire, yang2021ntire} and ``NTIRE 2020 challenge on real-world image SR"~\cite{lugmayr2020ntire} also represent the SOTA in this field.

\noindent In this paper, we propose Swin2SR, a SwinV2 Transformer-based model~\cite{liu2021swin, liu2022swin} for Compressed Image Super-Resolution and Restoration. This model represents a possible improvement or update of SwinIR~\cite{liang2021swinir} for these particular tasks. SwinV2~\cite{liu2022swin} (CVPR '22) allows us to tackle the major issues in training large transformer-based vision models, including training instability and duration, and resolution gaps between pre-training and fine-tuning~\cite{liang2021swinir}. 
We are the first work to explore successfully other transformer blocks beyond Swin Transformer~\cite{liu2021swin} for image super-resolution and restoration.
In some scenarios, our model can achieve similar results as SwinIR~\cite{liang2021swinir}, yet training 33\% less. 

We also provide extensive comparisons with state-of-the-art methods, and achieve competitive results at the related AIM 2022 Challenge.

\section{Related Work}
\label{sec:rel-work}

\subsection{Image Restoration}

Image restoration is split in a large number of sub-problems, for instance image denoising, image deblurring, super-resolution and compression artifacts removal among others.
Traditional model-based methods for image restoration were usually defined by hand-crafted priors that narrowed the ill-posed nature of the problems by reducing the set of plausible solutions~\cite{timofte2013anchored, timofte2014a, conde2022modelbased}.
Learning-based methods based on CNNs have recently gained great popularity for image restoration, and they represent current state-of-the-art in most low-level vision tasks (\ie{ denoising, deblurring, compression artifacts removal}). The first remarkable work on denoising with deep learning is probably Zhang \emph{et al.} \cite{zhang2017DnCNN} DnCNN. Other pioneering works include Dong \et{} SRCNN~\cite{dong2014srcnn} for image super-resolution and ARCNN~\cite{dong2015compression} for JPEG compression artifact removal.
Since research has moved towards deep learning, multiple CNN-based approaches have been proposed to improve the learned representations using more more complex neural network architectures, such as residual blocks, dense residual blocks, and laplacian operators~\cite{kim2016vdsr, cavigelli2017cas, zhang2021DPIR, zhang2018residualdense, zhang2020RDNIR, lai2017LapSRN}.
Other solutions attempt to exploit the attention mechanism in CNNs, such as channel attention and spatial attention~\cite{zhang2018rcan, dai2019SAN, niu2020HAN, liu2018NLRN, mei2021NLSA}.

\subsection{Vision Transformer}

The Transformer architecture~\cite{vaswani2017transformer} has recently gained much popularity in the computer vision community. Originally designed for neural machine translation, the Transformer architecture has successfully been applied to image classification~\cite{ dosovitskiy2020ViT, liu2021swin, vaswani2021SAhaloing, conde2021exploring, conde2021clip}, object detection~\cite{carion2020DETR, touvron2020DeiT}, object segmentation~\cite{cao2021swinunet} and perceptual quality assessment (IQA)~\cite{conde2022conformer, gu2022ntire}. The attention mechanism learns complex global interactions by attending to important regions in the image. Due to its impressive performance, transformers have also been introduced to image restoration~\cite{chen2021IPT, cao2021videosr, wang2021uformer}. 
More recently, Chen~\et~\cite{chen2021IPT} proposed IPT, a general backbone model for multiple image restoration tasks based on the standard Transformer~\cite{vaswani2017transformer}. This model shows promising performance on several tasks, however, it relies on a large number of parameters and heavy computation (over 115.5M parameters), and a large-scale dataset like ImageNet (over 1M images). VSR-Transformer proposed by Cao~\et~\cite{cao2021videosr} combines the self-attention mechanism and CNN-based feature extraction to fuse better features in video super-resolution. Note that many transformer-based approaches such as IPT~\cite{chen2021IPT} and VSR-Transformer~\cite{cao2021videosr} use patch-wise attention, which may not be optimal for image restoration. Liang \et{} proposed SwinIR~\cite{liang2021swinir} based Swin Transformer~\cite{liu2021swin}, which represents the state-of-the-art in many restoration tasks.
%

In this context, the Swin Transformer~\cite{liu2021swin} improved the Vision Transformer architecture by using shifted window based self-attention with progressive image downsampling like CNNs. Window self-attention is computed for non-overlapped image patches reducing attention computational complexity from Eq.~\ref{eq:swin-attn} to Eq.~\ref{eq:swin2-attn}:
\begin{equation}
    O(MSA) = 4hwC^{2} + 2(hw)^{2}C
\label{eq:swin-attn}
\end{equation}

\begin{equation}
    O(WMSA) = 4hwC^{2} + 2M^{2}hwC
\label{eq:swin2-attn}
\end{equation}
for an image of size $h \times w$ and patches of size $M \times M$. The former quadratic computational complexity is replaced by a linear complexity when $M$ is fixed. Learned relative positional bias are also added to include position information while computing similarities for each head.

The Swin Transformer V2~\cite{liu2022swin} modified the Swin Attention~\cite{liu2021swin} module to better scale model capacity and window resolution. They first replace the \textit{pre-norm} by a \textit{post-norm} configuration, use a \textit{scaled cosine attention} instead of the \textit{dot product attention} and use a \textit{log-spaced continuous} relative position bias approach to replace the previous \textit{parameterized} approach. The attention output is:
\begin{equation}
    Attention(Q, K, V ) = Softmax(cos(Q, K) / \tau + S)V
\end{equation}
Where $Q,K,V \in \mathbb{R}^{M^{2} \times d}$ are the query, key and value matrices. $S \in \mathbb{R}^{M^{2} \times M^{2}}$ are the relative to absolute positional embeddings obtained by projecting the position bias after re-indexing. $\tau$ is a learnable scalar, non-shared across heads and layers. This block is illustrated in Figure~\ref{fig:main}. 

\section{Our Method}
\label{sec:ours}

\begin{figure}
    \centering
    \includegraphics[width=\textwidth]{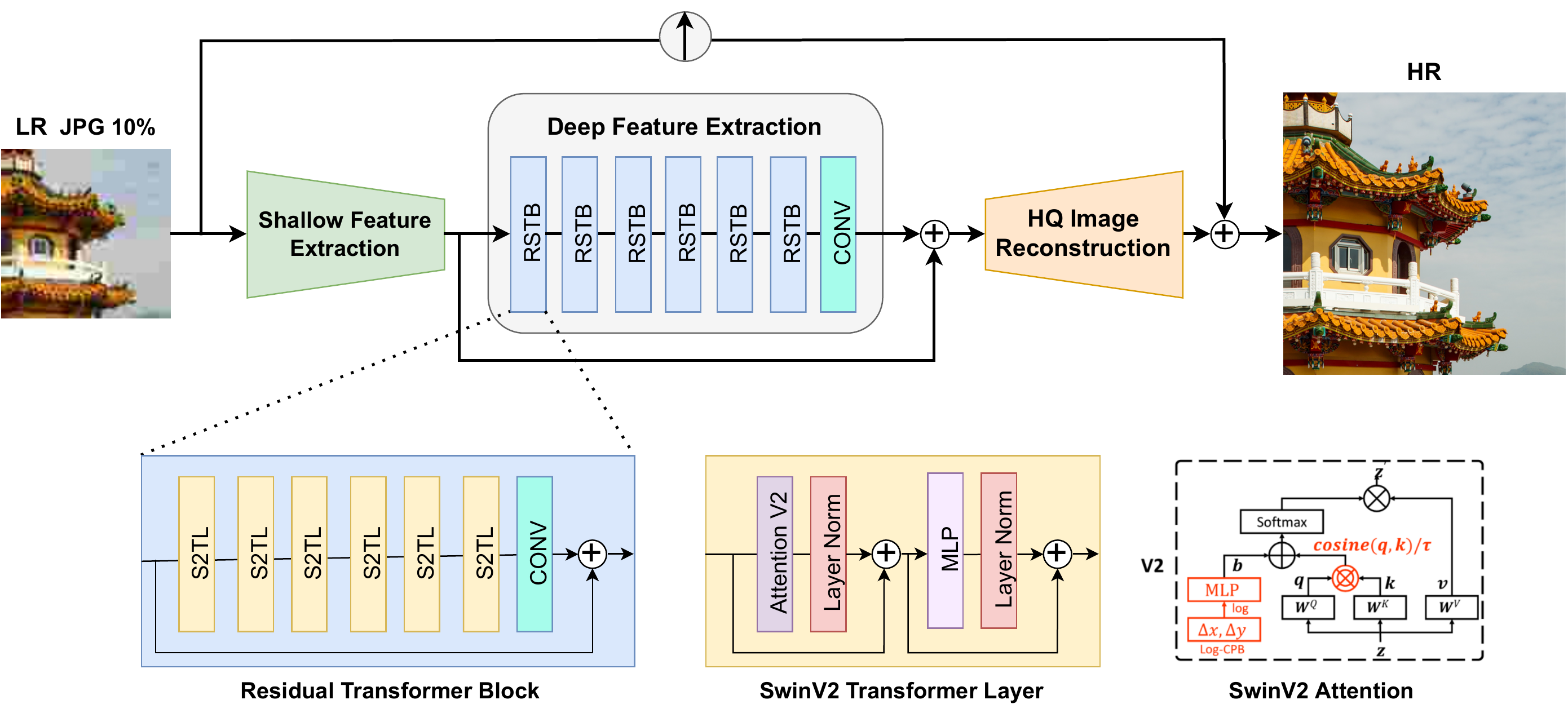}
    \caption{The architecture of the proposed Swin2SR~\cite{conde2022swin2sr}. In this case, we show our method applied to Super-Resolution of Compressed Image~\cite{yang2022aim}.}
    \label{fig:main}
\end{figure}

Our method \ours is illustrated in Figure~\ref{fig:main}. We propose some modifications of SwinIR~\cite{liang2021swinir}, which is based on Swin Transformer~\cite{liu2021swin}, that enhance the model's capabilities for Super-Resolution, and in particular, for Compressed Input SR. We update the original Residual Transformer Block (RSTB) by using the new SwinV2 transformer~\cite{liu2022swin} (CVPR'22) layers and attention to scale up capacity and resolution~\cite{liu2022swin}.
Our method has a classical upscaling branch which uses a bicubic interpolation, as shown in the AIM 2022 Challenge Leaderboard~\cite{yang2022aim} and our results~(\ref{tab:aim-results}), this alone can recover basic structural information. For this reason, the output of our model is added to the basic upscaled image, to enhance it.
We also explore different loss functions to make our model more robust to JPEG compression artifacts, being able to recover high-frequency details from the compressed LR image, and therefore, achieve better performance.

\paragraph{\textbf{Advantages of updating to SwinV2}}
The SwinV2 architecture modifies the shifted window self-attention module to better scale model capacity and window resolution. The use of \textit{post normalization} instead of \textit{pre normalization} reduce the average feature variance of deeper layers and increase numerical stability during training. This allows to scale the SwinV2 Transformer up to 3 billion parameters without training instabilities~\cite{liu2022swin}. The use of \textit{scaled cosine attention} instead of \textit{dot product} between queries and keys reduce the dominance of some attention heads for a few pixel pairs. In some tasks, our \ours model achieved the same results as SwinIR~\cite{liang2021swinir}, yet training 33\% less iterations. Finally, the use of \textit{log-spaced continuous} relative position bias allows us to generalize to higher input resolution at inference time. 

\subsection{Experimental Setup}

For a fair comparison and ensure reproducibility, we follow the same experimental setup as SwinIR~\cite{liang2021swinir} and other state-of-the-art methods~\cite{kai2021bsrgan, zhang2018residualdense}. 

We evaluate our model on three tasks: JPEG compression artifacts removal (Section~\ref{sec:jpeg}), classical and lightweight image super-resolution (Section~\ref{sec:classical}) and compressed image super-resolution (Section~\ref{sec:aim-challenge}).
We mainly use the DIV2K dataset for training and validation~\cite{agustsson2017ntire}, and following the tradition of image SR, we report PSNR and SSIM on the Y channel of the YCbCr space~\cite{liang2021swinir, kai2021bsrgan, zhang2018residualdense}.

Our model \ours has the following elements, similar to SwinIR~\cite{liang2021swinir}: shallow feature extraction, deep feature extraction and high-quality image reconstruction modules. 
The \textbf{shallow feature extraction} module uses a convolution layer to extract features, which are directly transmitted to the reconstruction module to preserve low-frequency information~\cite{zhang2017DnCNN, liang2021swinir}. The \textbf{Deep feature extraction} module is mainly composed of Residual SwinV2 Transformer blocks (RSTB), each of which utilizes several SwinV2 Transformer~\cite{liu2022swin} layers (S2TL) for local attention and cross-window interaction. 
Finally, both shallow and deep features are fused in the reconstruction module for high-quality image reconstruction. To upscale the image, we use standard a pixel shuffle operation.

The hyper-parameters of the architecture are as follows: the RSTB number, S2TL number, window size, channel number and attention head number are generally set to 6, 6, 8, 180 and 6, respectively. For lightweight image SR, we explain the details in Section~\ref{sec:classical}.

\subsection{Implementation details}

The method was implemented in Pytorch using as baseline \url{https://github.com/cszn/KAIR} and the official repository for SwinIR~\cite{liang2021swinir}.
We initially train \ours from scratch using the basic $\mathcal{L}_1$ loss for reconstruction. While training, we randomly crop HR images using 192px patch size and crop correspondingly the LR image generated offline using MATLAB, we also use standard augmentations that include all variations of flipping and rotations~\cite{timofte2016seven}.
We use mainly the DIV2K~\cite{agustsson2017ntire}. In some experiments, to explore the potential benefits of more training data, we also use the Flickr2K dataset (2650 images).

\begin{figure}
    \centering
    \includegraphics[width=\textwidth]{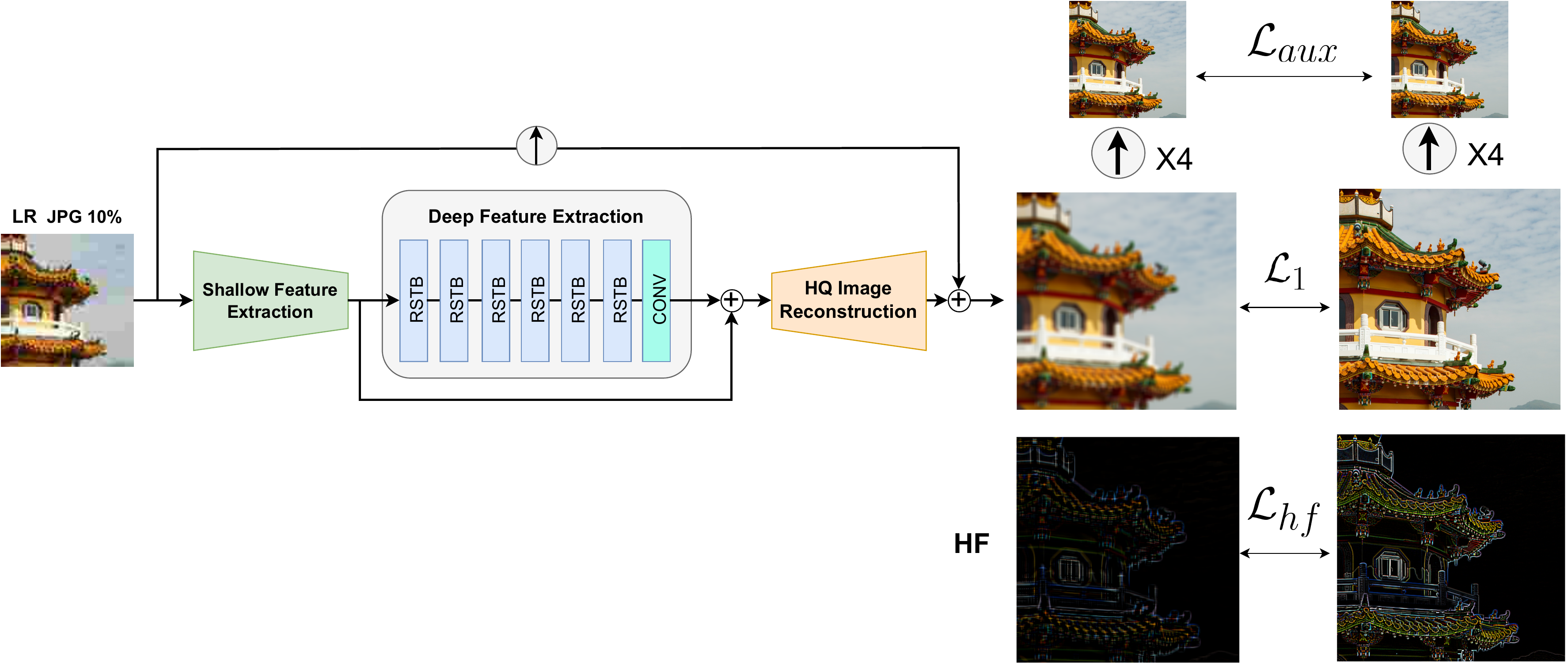}
    \caption{Swin2SR training with additional regularization.}
    \label{fig:loss}
\end{figure}

In the particular scenario of \textbf{Compressed Input Super-Resolution}~\cite{yang2022aim} (Section~\ref{sec:aim-challenge}), we explore different loss functions to improve the performance and robustness of our method; these are represented in Figure~\ref{fig:loss}.

First, we add an Auxiliary Loss that minimizes the $\mathcal{L}_1$ distance between the downsampled prediction $\hat{y}$ and the downsampled reference $y$ \texttt{.png}, as follows:

\begin{equation}
\label{eq:auxlos}
\mathcal{L}_{aux} = \left \| D(y) - D(\hat{y}) \right \|_{1}
\end{equation}

where $x$ is the low-resolution degraded image, $y$ is the high-resolution clean image, $f(x)=\hat{y}$ is the restored image using our model $f$, and $D(.)$ is a downsampling operator (\ie{} $\times$4 bicubic kernel).
This helps to ensure consistency also at lower-resolution. In order to minimize Eq.~\ref{eq:auxlos} the restored image at a lower resolution should not have artifacts (i.e. the prediction at lower resolution should be close to the downsampled reference \texttt{.png} without artifacts).

Second, we extract the high-frequency (HF) information from the High-Resolution images. This loss is formulated as follows:

\begin{equation}
\label{eq:hfloss}
\mathcal{L}_{hf} = \left \| (y - (y*b)) - (\hat{y} - (\hat{y}*b))  \right \|_{1} = 
\left \| HF(y) - HF(\hat{y})  \right \|_{1}
\end{equation}

where $HR(.)$ denotes the high-frequency information of an image. To obtain this, we convolve a simple $5\times5$ kernel $b$ as a gaussian blur operation. This term enforces the prediction to have the same high-frequency details as the reference, and therefore, it helps to improve the sharpness and quality of the results.

\section{Experimental Results}
\label{sec:exp}

\subsection{JPEG Compression Artifacts Removal}
\label{sec:jpeg}

Table~\ref{tab:jpeg-results} shows the comparison of \ours with \textit{state-of-the-art} JPEG compression artifact reduction methods: ARCNN~\cite{dong2015compression}, DnCNN-3~\cite{zhang2017DnCNN}, QGAC~\cite{ehrlich2020quantization}, RNAN~\cite{zhang2019RNAN}, and MWCNN~\cite{liu2018multi}. All of compared methods are CNN-based models trained especifically for each quality type (\ie{ four models per dataset}). Due to our limited resources, and seeking for a more flexible approach, we train a single model able to deal with the four different quality factors. For this reason, we do not compare directly with DRUNet~\cite{zhang2021DPIR}, as we consider it an unfair comparison. Moreover, \ours only has 12M parameters, while DRUNet~\cite{zhang2021DPIR}, is a large model that has 32.7M parameters. Note that we perform these comparisons using the same setup as~\cite{liang2021swinir}.
Following~\cite{zhang2020RDNIR, zhang2021DPIR, liang2021swinir}, we test different methods on two benchmark datasets: (i) Classic5~\cite{foi2007Classic5} and (ii) LIVE1~\cite{sheikh2005live}; using JPEG quality factors ($q$) 10, 20, 30 and 40. 
As we can see in Table~\ref{tab:jpeg-results}, our \ours achieves state-of-the-art results in compression artifacts removal.

\vspace{-5mm}

\begin{table}[!h]
\begin{center}
\caption{Quantitative comparison (average PSNR/SSIM) with state-of-the-art methods for \textbf{JPEG compression artifact reduction} on benchmark datasets. Best and second best performance are in \R{red} and \B{blue} colors, respectively. Note that \ours is a single model that generalizes to different qualities, meanwhile, some methods are trained for each specific quality. Some numbers are from \cite{jiang2021towards}.}
\label{tab:jpeg-results}
\adjustbox{max width=\textwidth}{%
\begin{tabular}{c|c|c|c|c|c|c|c|c}
\hline
Dataset & $q$
& ARCNN~\cite{dong2015compression} 
& DnCNN~\cite{zhang2017DnCNN} 
& QGAC~\cite{ehrlich2020quantization}
& RNAN~\cite{zhang2019RNAN}
& MWCNN~\cite{liu2018multi}
& SwinIR~\cite{liang2021swinir}
& \textbf{\ours}\\
\hline
\hline
\multirow{4}{*}{\makecell{Classic5\\\cite{foi2007Classic5}}} & 10
& 29.03/0.79
& 29.40/0.80
& 29.84/0.83
& 29.96/0.81
& 30.01/0.82
& \R{30.27/0.82}
& \B{30.02/0.81}
\\
& 20
& 31.15/0.85
& 31.63/0.86
& 31.98/0.88
& 32.11/0.86
& 32.16/\B{0.87}
& \B{31.32}/0.85
& \R{32.26/0.87}
\\
& 30
& 32.51/0.88
& 32.91/0.88
& 33.22/0.90
& 33.38/0.89
& \B{33.43/0.89}
& {31.39/0.853}
& \R{33.51/0.89}
\\
& 40
& 33.32/0.89
& 33.77/0.90
& -
& 34.27/0.90
& \B{34.27/0.90}
& 31.38/0.85
& \R{34.33/0.90}
\\
\hline
\hline
\multirow{4}{*}{\makecell{LIVE1\\\cite{sheikh2005live}}} & 10
& 28.96/0.80
& 29.19/0.81
& 29.53/0.84
& 29.63/0.82
& \B{29.69/0.82}
& \R{29.86/0.82}
& 29.67/\R{0.82}
\\
& 20
& 31.29/0.87
& 31.59/0.88
& 31.86/0.90
& 32.03/0.88
& \B{32.04/}\B{0.89}
& {31.00/0.86}
& \R{32.07/}\R{0.89}
\\
& 30
& 32.67/0.90
& 32.98/0.90
& 33.23/0.92
& 33.45/0.91
& \B{33.45/0.91}
& {31.08/0.86}
& \R{33.49/0.91}
\\
& 40
& 33.63/0.91
& 33.96/0.92
& -
& 34.47/0.92
& \B{34.45}/\R{0.93}
& {31.05/0.86}
& \R{34.49}/\B{0.92}
\\
\hline
\end{tabular}}
\end{center}
\end{table}

\begin{figure}[!ht]
    \centering
    \setlength{\tabcolsep}{2.0pt}
    \begin{tabular}{cccc}
    \includegraphics[width=0.24\linewidth]{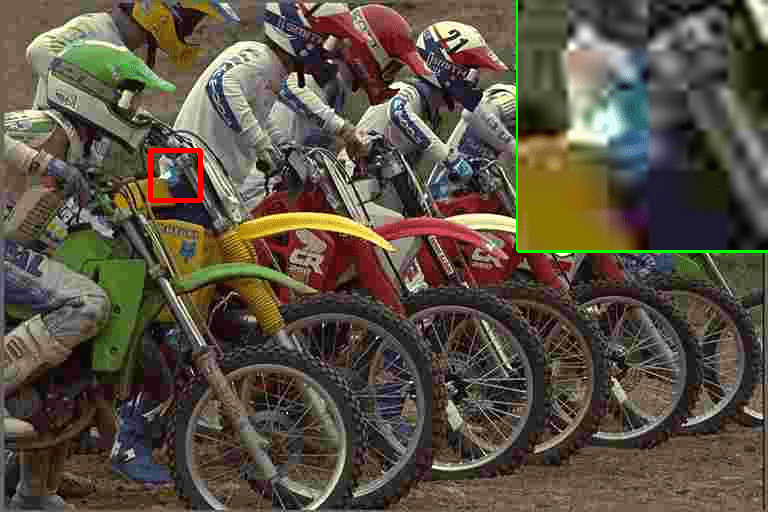} &
    \includegraphics[width=0.24\linewidth]{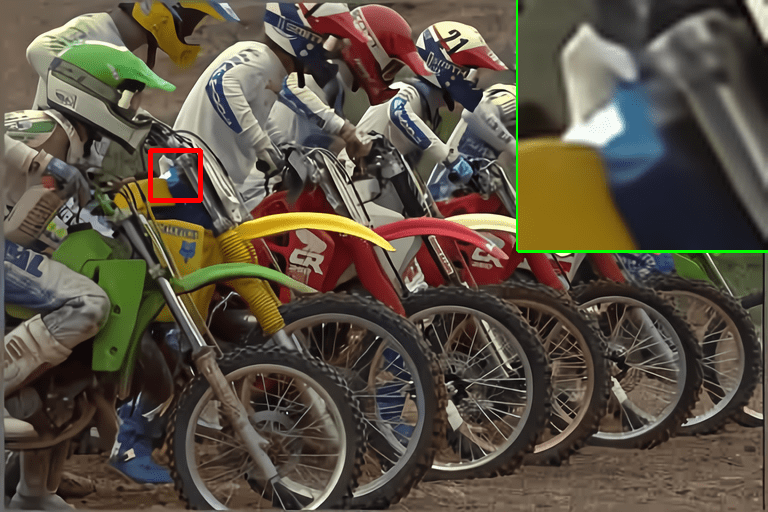} &    \includegraphics[width=0.24\linewidth]{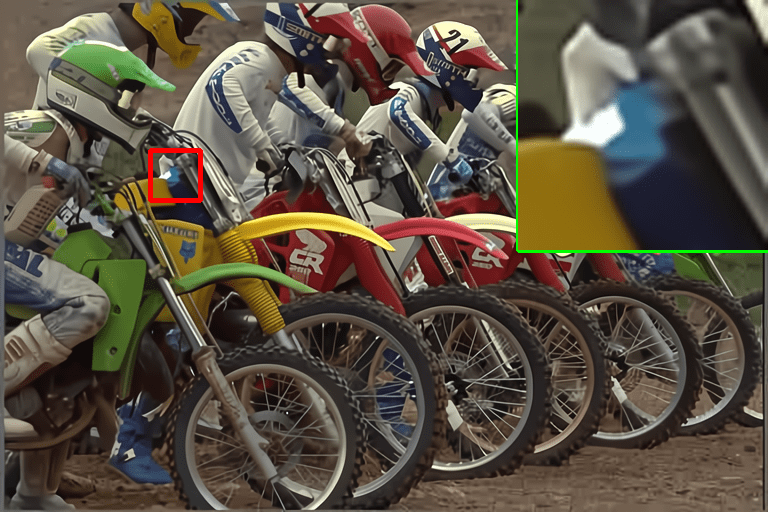} & 
    \includegraphics[width=0.24\linewidth]{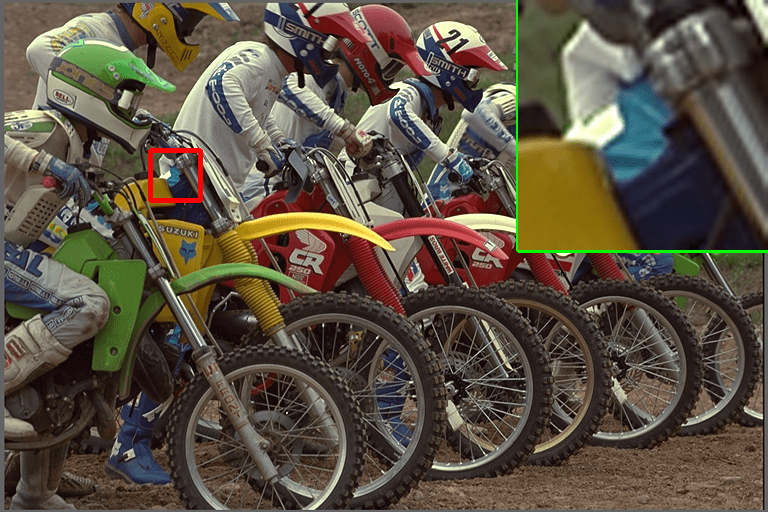}
    \tabularnewline
    \includegraphics[width=0.24\linewidth]{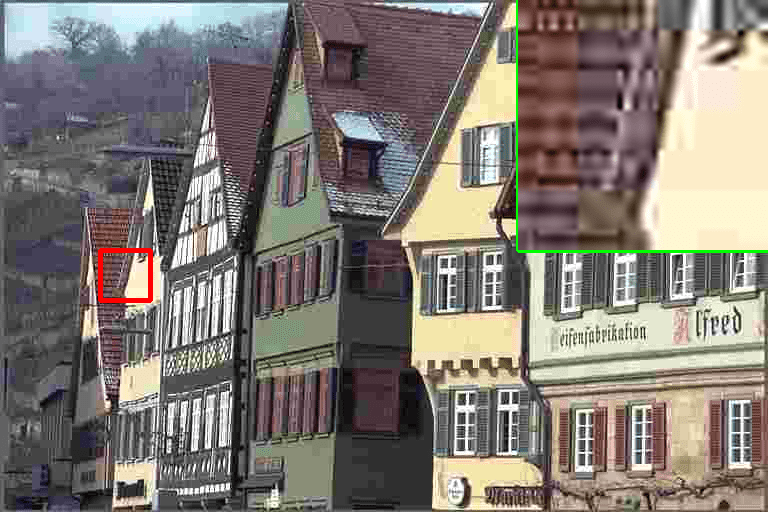} &
    \includegraphics[width=0.24\linewidth]{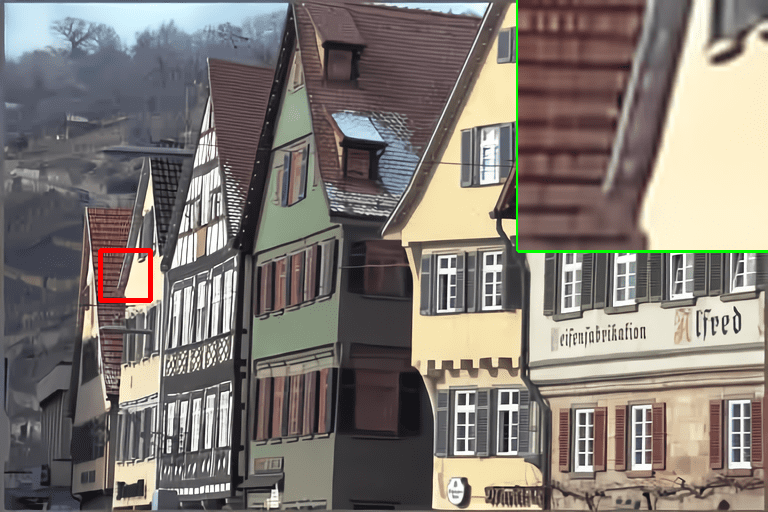} &
    \includegraphics[width=0.24\linewidth]{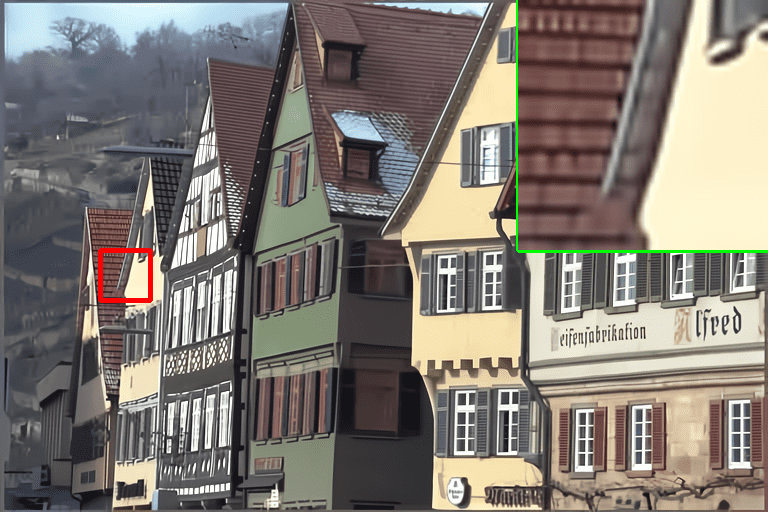} & 
    \includegraphics[width=0.24\linewidth]{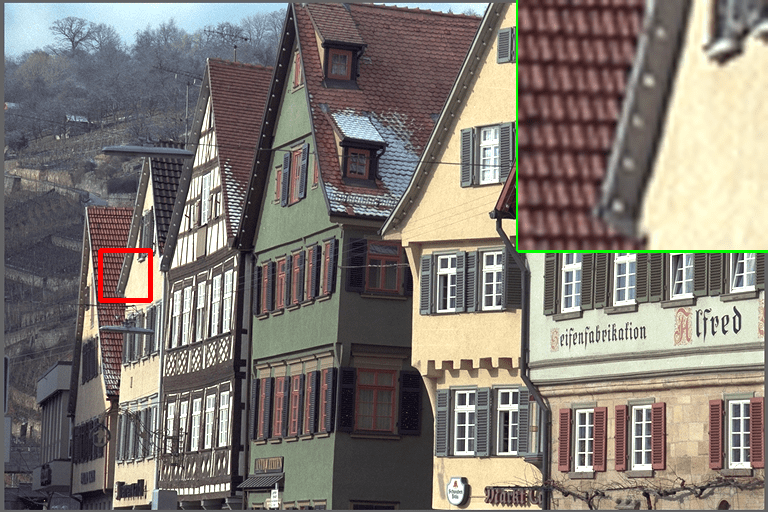}
    \tabularnewline
    \includegraphics[width=0.24\linewidth]{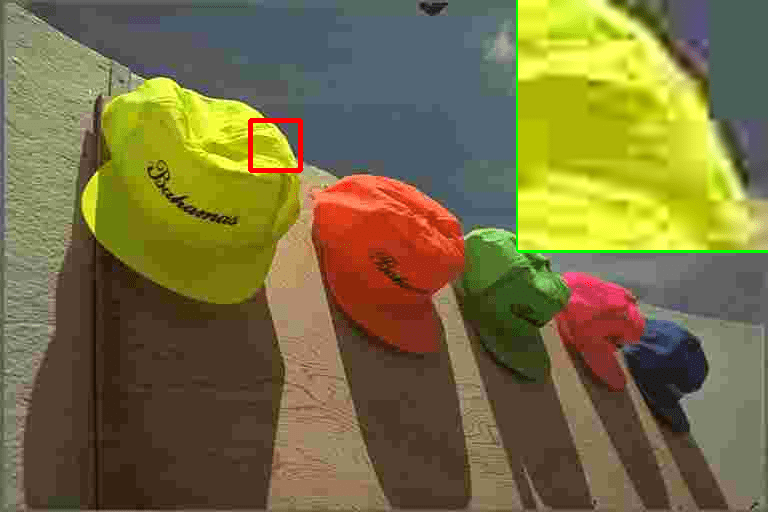} &
    \includegraphics[width=0.24\linewidth]{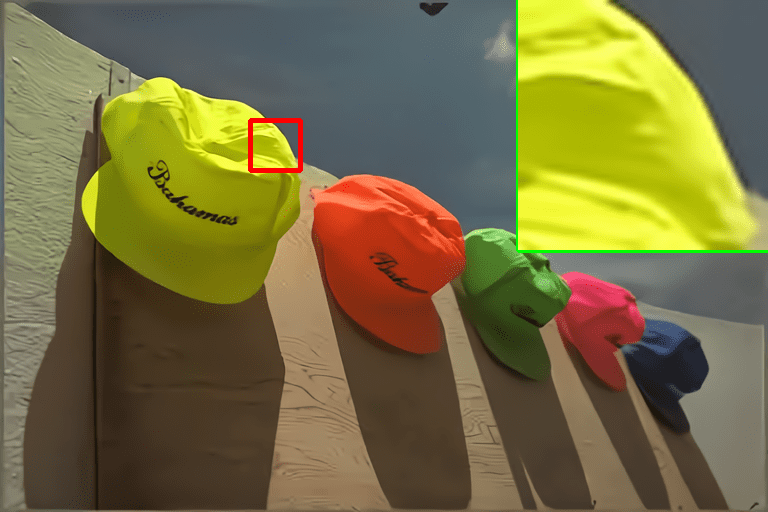} &
    \includegraphics[width=0.24\linewidth]{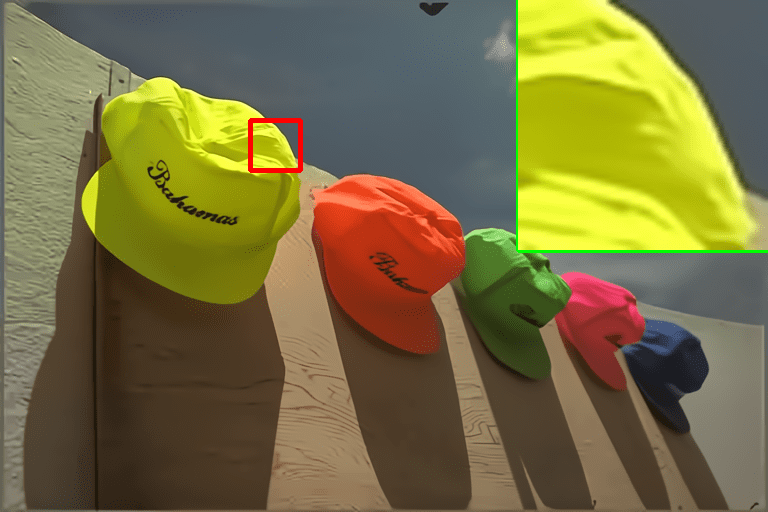} & 
    \includegraphics[width=0.24\linewidth]{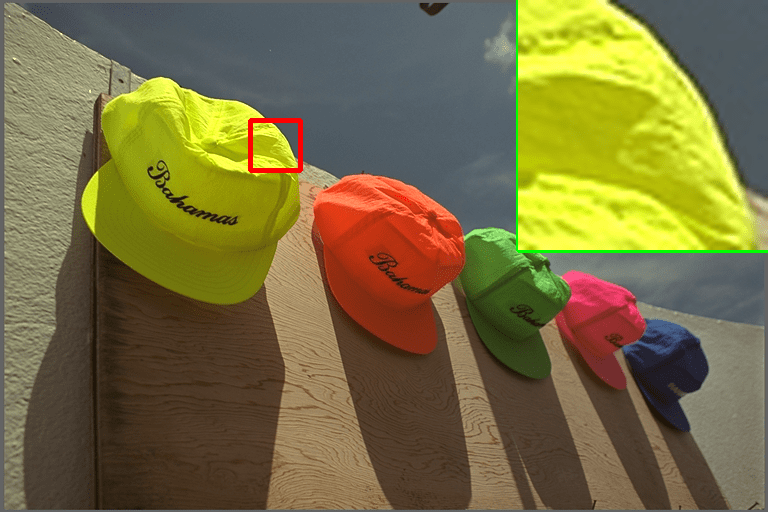}
    \tabularnewline
    Input Compressed & FBCNN ($q=10$)~\cite{liang2021swinir} & Swin2SR (ours) & Reference
    \tabularnewline
    \end{tabular}
    \caption{Qualitative samples of JPEG Compression Artifacts Removal. We show the JPEG compressed image at quality $q=10$. All images have the same resolution. Images from Classic5~\cite{foi2007Classic5} and LIVE1~\cite{sheikh2005live}. Best viewed by zooming.}
    \label{fig:jpeg-results}
\end{figure}

In the case of SwinIR~\cite{liang2021swinir}, which is also \textit{state-of-the-art} for JPEG artifacts reduction, authors train \underline{one model per quality factor} (\ie four models) for 1600K iterations, and $q=10/20/30$ models are fine-tuned using the $q=40$ model as general baseline. We train a \underline{single model} using the same setup~\cite{liang2021swinir}, only for 800k iterations (\ie{ $\times2$ less training than SwinIR~\cite{liang2021swinir}}), and JPEG compression as an augmentation.  For this reason in Table~\ref{tab:jpeg-results} we compare with SwinIR trained for the most challenging $q=10$.
We also compare with MWCNN~\cite{liu2018multi}, IDCN~\cite{zheng2019implicit} and FBCNN-C~\cite{jiang2021towards} using RGB color images.
Attending to Tables~\ref{tab:jpeg-results}~and~\ref{tab:jpeg-comp-swinir}, we consider our model a more general and flexible approach for grayscale or color compression artifacts removal, since it can be trained faster and generalizes to different compression quality factors.
We also provide \textbf{qualitative results} in Figure~\ref{fig:jpeg-results}. \ours can restore compressed images and generate high-quality results. We provide additional results in the supplementary material.


\begin{table}[!h]
\begin{center}
\caption{Quantitative comparison on \textbf{color} JPEG images with \textbf{single} compression. We report average PSNR/SSIM on benchmark datasets. Our model outperforms networks designed for this particular task (although we recognise that training with more data). Some numbers are from~\cite{jiang2021towards}.}
\label{tab:jpeg-comp-swinir}
\adjustbox{max width=\textwidth}{%
\begin{tabular}{c|c|c|c|c|c|c|c|c}

\hline
Dataset & $q$
& JPEG
& ARCNN~\cite{dong2015compression} 
& QGAC~\cite{ehrlich2020quantization}
& MWCNN~\cite{liu2018multi}
& IDCN~\cite{zheng2019implicit}
& FBCNN-C~\cite{jiang2021towards}
& \textbf{\ours}\\
\hline
\hline
\multirow{2}{*}{\makecell{LIVE1\\\cite{sheikh2005live}}} & 10
& 25.69/0.74
& 26.91/0.79
& 27.62/0.80
& 27.45/0.80
& 27.63/0.81
& \B{27.77/0.80}
& \R{27.98/0.82}
\\
& 40
& 30.28/0.88
& -
& 32.05/0.91
& -
& -
& \B{32.34/0.91}
& \R{32.53/0.92}
\\
\hline
\hline
\multirow{2}{*}{\makecell{ICB\\\cite{icb}}} & 10
& 29.44/0.75
& 30.06/0.77
& 32.06/0.81
& 30.76/0.77
& 31.71/0.80
& \B{32.18/0.81}
& \R{32.46/0.81}
\\
& 40
& 33.95/0.84
& -
& 32.25/0.91
& -
& -
& \B{36.02/0.86}
& \R{36.25/0.86}
\\
\hline
\end{tabular}}
\end{center}
\end{table}

\subsection{Classical Image Super-Resolution}
\label{sec:classical}

\begin{table}[!ht]
\begin{center}
\caption{Quantitative comparison (average PSNR/SSIM) with state-of-the-art methods for \textbf{classical image SR} on benchmark datasets. Best and second best performance are in \R{red} and \B{blue} colors, respectively.}%
\label{tab:sr_results}
\vspace{2.5mm}
\resizebox{\linewidth}{!}{%
\begin{tabular}{l|c|c|c|c|c|c|c|c|c|c|c|c}
\hline
\multirow{2}{*}{Method} & \multirow{2}{*}{Scale} & \multirow{2}{*}{\makecell{Training\\Dataset}} &  \multicolumn{2}{c|}{Set5~\cite{Set5}} &  \multicolumn{2}{c|}{Set14~\cite{Set14}} &  \multicolumn{2}{c|}{BSD100~\cite{BSD100}} &  \multicolumn{2}{c|}{Urban100~\cite{huang2015single}} &  \multicolumn{2}{c}{Manga109~\cite{Manga109}} 
\\
\cline{4-13}
&  &  & PSNR & SSIM & PSNR & SSIM & PSNR & SSIM & PSNR & SSIM & PSNR & SSIM 
\\
\hline
\hline
RCAN~\cite{zhang2018rcan} & $\times$2 & DIV2K %
& {38.27}
& {0.9614}
& {34.12}
& {0.9216}
& {32.41}
& {0.9027}
& {33.34}
& {0.9384}
& {39.44}
& {0.9786}
\\  
SAN~\cite{dai2019SAN} & $\times$2 & DIV2K %
& {38.31}
& {0.9620}
& {34.07}
& {0.9213}
& {32.42}
& {0.9028}
& {33.10}
& {0.9370}
& {39.32}
& {0.9792}
\\
IGNN~\cite{zhou2020IGNN} & $\times$2 & DIV2K %
& {38.24}
& {0.9613}
& {34.07}
& {0.9217}
& {32.41}
& {0.9025}
& {33.23}
& {0.9383}
& {39.35}
& {0.9786}
\\
HAN~\cite{niu2020HAN} & $\times$2 & DIV2K %
& {38.27}
& {0.9614}
& {34.16}
& {0.9217}
& {32.41}
& {0.9027}
& {33.35}
& {0.9385}
& {39.46}
& {0.9785}  
\\ 
NLSA~\cite{mei2021NLSA} & $\times$2 & DIV2K %
& 38.34 
& 0.9618 
& 34.08 
& 0.9231
& 32.43 
& 0.9027 
& 33.42
& 0.9394
& 39.59
& 0.9789
\\
\hdashline
DBPN~\cite{haris2018DBPN} & $\times$2 & DIV2K+Flickr2K%
& 38.09
& 0.9600
& 33.85
& 0.9190
& 32.27
& 0.9000
& 32.55
& 0.9324
& 38.89
& 0.9775        
\\
IPT~\cite{chen2021IPT} & $\times$2  & ImageNet%
& {38.37}
& {-}
& {34.43}
& {-}
& {32.48}
& {-}
& {33.76}
& {-}
& {-}
& {-}
\\
SwinIR~\cite{liang2021swinir} & $\times$2  & DIV2K+Flickr2K
& \B{38.42}
& \B{0.9623}
& \B{34.46}
& \B{0.9250}
& \B{32.53}
& \B{0.9041}
& \B{33.81}
& \B{0.9427}
& \R{39.92}
& \B{0.9797}
\\
\rowcolor{Gray}\textbf{\algname{}} & $\times$2  & DIV2K+Flickr2K
& \R{38.43}
& \R{0.9623}
& \R{34.48}
& \R{0.9256}
& \R{32.54}
& \R{0.905}
& \R{33.89}
& \R{0.9431}
& \B{39.88}
& \R{0.9798}
\\
\rowcolor{Gray}\textbf{Swin2SR-D} & $\times$2  & DIV2K+Flickr2K
& {38.06}
& {-}
& {33.81}
& {-}
& {32.32}
& {-}
& {32.6}
& {-}
& {38.98}
& {-}
\\
\hline
\hline
RCAN~\cite{zhang2018rcan}& $\times$4  & DIV2K
& {32.63}
& {0.9002}
& {28.87}
&{0.7889}
& {27.77}
& {0.7436}
&{26.82}
& {0.8087}
&{31.22}
& {0.9173}
\\ 
SAN~\cite{dai2019SAN} & $\times$4  & DIV2K
& {32.64}
& {0.9003}
& {28.92}
& {0.7888}
& {27.78}
& {0.7436}
& {26.79}
& {0.8068}
& {31.18}
& {0.9169}
\\
IGNN~\cite{zhou2020IGNN}  & $\times$4  & DIV2K
& {32.57}
& {0.8998}
& {28.85}
& {0.7891}
& {27.77}
& {0.7434}
& {26.84}
& {0.8090}
& {31.28}
& {0.9182}
\\
HAN~\cite{niu2020HAN}  & $\times$4  & DIV2K
& {32.64}
& {0.9002}
& {28.90}
& {0.7890}
& {27.80}
& {0.7442}
& {26.85}
& {0.8094}
& {31.42}
& {0.9177}
\\
NLSA~\cite{mei2021NLSA} & $\times$4 & DIV2K
& 32.59 
& 0.9000 
& 28.87 
& 0.7891 
& 27.78 
& 0.7444 
& {26.96}
& {0.8109}
& 31.27 
& 0.9184
\\
\hdashline
DBPN~\cite{haris2018DBPN} & $\times$4 & DIV2K+Flickr2K
& 32.47
& 0.8980
& 28.82
& 0.7860
& 27.72
& 0.7400
& 26.38
& 0.7946
& 30.91
& 0.9137
\\
IPT~\cite{chen2021IPT} & $\times$4 & ImageNet
& {32.64}
& {-}
& {29.01}
& {-}
& {27.82}
& {-}
& {27.26}
& {-}
& {-}
& {-}
\\
RRDB~\cite{wang2018esrgan} & $\times$4 & DIV2K+Flickr2K
& {32.73}
& {0.9011 }
& {28.99}
& {0.7917}
& {27.85}
& {0.7455}
& {27.03}
& {0.8153}
& {31.66}
& {0.9196}
\\
SwinIR~\cite{liang2021swinir}  & $\times$4  & DIV2K+Flickr2K
& \R{32.92}
& \R{0.9044}
& \R{29.09}
& \R{0.7950}
& \B{27.92}
& \B{0.7489}
& \B{27.45}
& \B{0.8254}
& \R{32.03}
& \R{0.9260}
\\
\rowcolor{Gray}\textbf{\algname{}} & $\times$4  & DIV2K+Flickr2K
& \R{32.92}
& \B{0.9039}
& \B{29.06}
& \B{0.7946}
& \R{27.92}
& \R{0.7505}
& \R{27.51}
& \R{0.8271}
& \B{31.03}
& \B{0.9256}
\\
\rowcolor{Gray}\textbf{Swin2SR-D} & $\times$4  & DIV2K+Flickr2K
& {32.41}
& {-}
& {28.75}
& {-}
& {27.69}
& {-}
& {26.4}
& {-}
& {30.96}
& {-}
\\
\hline             
\end{tabular}}
\end{center}
\end{table}

For classical and lightweight image SR, following~\cite{liang2021swinir, kai2021bsrgan,zhang2021DPIR}, we train \ours on 800 training images of DIV2K and 2650 images from Flickr2K. For fair comparison with SwinIR~\cite{liang2021swinir}, we use $64\times64$ LQ image patches, and the HQ-LQ image pairs are obtained by the MATLAB bicubic kernel.
We train our model from scratch during 500k iterations, and fine-tune it for the $\times4$ task.
Table~\ref{tab:sr_results} shows the quantitative comparisons between \ours and \textit{state-of-the-art methods}: DBPN~\cite{haris2018DBPN}, RCAN~\cite{zhang2018rcan}, RRDB~\cite{wang2018esrgan}, SAN~\cite{dai2019SAN}, IGNN~\cite{zhou2020IGNN}, HAN~\cite{niu2020HAN}, NLSA~\cite{mei2021NLSA}, IPT~\cite{chen2021IPT} and SwinIR~\cite{liang2021swinir}. 
All the CNN-based methods perform worse than the studied transformer-based methods, IPT~\cite{chen2021IPT}, SwinIR~\cite{liang2021swinir} and Swin2SR. Moreover, \ours was trained using only DIV2K+Flickr2K and achieves better performance than IPT~\cite{chen2021IPT}, even though IPT~\cite{chen2021IPT} utilizes ImageNet (more than 1.3M images) in training and has huge number of parameters (115.5M). In contrast, \ours has only 12M parameters, which is competitive even compared with state-of-the-art CNN-based models (15.4$\sim$44.3M).
Note that our models achieve essentially the same performance as SwinIR~\cite{liang2021swinir}, yet trained for 400k iterations from scratch, without fine-tuning or pre-training, in comparison with SwinIR~\cite{liang2021swinir} models trained during 500k, and in the case of $\times4$ fine-tuned using the $\times2$ model.
We provide visual comparisons in Figures~\ref{fig:aim-results}.
\ours can remove artifacts and recover structural information and high-frequency details.

\paragraph{\textbf{Dynamic Super-Resolution}}
Likewise Section~\ref{sec:jpeg}, we explore the performance of a single super-resolution model to upscale directly using any arbitrary $\times$ factor. We call this a Dynamic Super-Resolution model, referred as Swin2SR-D. 

In SwinIR~\cite{liang2021swinir} we can find an upsampling layer designed to upscale images using s particular factor (\ie{ $\times2$}). This layer cannot be adjusted to a different factor on-line, therefore, SwinIR~\cite{liang2021swinir} trains one model for each different factor. To deal with this problem, we implemented a Dynamic upsampling layer, which initially can super-resolve the images using $\times2$, $\times3$, and $\times4$ factors on-line in the same module. 
We show in Table~\ref{tab:sr_results} the potential of this method, as this single model can perform $\times2$ and $\times4$ super-resolution indistinctly.

\paragraph{\textbf{Lightweight image SR.}} We also provide comparison of Swin2SR-s with \textit{state-of-the-art methods} lightweight image SR methods: CARN~\cite{ahn2018CARN}, FALSR-A~\cite{chu2021fast}, IMDN~\cite{hui2019imdn}, LAPAR-A~\cite{li2021lapar}, LatticeNet~\cite{luo2020latticenet} and SwinIR (small)~\cite{liang2021swinir}.

Our lightweight model is designed as SwinIR (small)~\cite{liang2021swinir}, we decrease the number of Residual Swin Transformer Blocks (RSTB) and convolution channels to 4 and 60, respectively. However, the number of Swin Transformer Layers (STL) in each RSTB, window size and attention head number still set to 6, 8 and 6, respectively (as in \ours base model).

In addition to PSNR and SSIM, we also report the total numbers of parameters and multiply-accumulate operations for different methods~\cite{liang2021swinir}. These MACs are calculated using a $1280\times 720$ image. As shown in Table~\ref{tab:lightweight_sr_results}, \ours outperforms competitive methods~\cite{ahn2018CARN, chu2021fast, hui2019imdn, li2021lapar} on different benchmark datasets, with similar total numbers of parameters and multiply-accumulate operations.
In our experiments, \ours can achieve the same results as SwinIR (small)~\cite{liang2021swinir}, yet, training almost 33\% less iterations.

\vspace{-5mm}

\begin{table}[!h]
\scriptsize
\begin{center}
\caption{Quantitative comparison (average PSNR/SSIM) with state-of-the-art methods for \textbf{lightweight image SR $\times$2} on benchmark datasets. Best and second best performance are in \R{red} and \B{blue} colors, respectively. In our experiments, Swin2SR-s converges faster than SwinIR (small)~\cite{liang2021swinir}.}
\vspace{2.5mm}
\label{tab:lightweight_sr_results}
\resizebox{\textwidth}{!}{
\begin{tabular}{l|c|c|c|c|c|c|c|c|c|c|c|c|c}
\hline
\multirow{2}{*}{Method} & \multirow{2}{*}{\#~Params} & \multirow{2}{*}{\# Mult-Adds} &  \multicolumn{2}{c|}{Set5~\cite{Set5}} &  \multicolumn{2}{c|}{Set14~\cite{Set14}} &  \multicolumn{2}{c|}{BSD100~\cite{BSD100}} &  \multicolumn{2}{c|}{Urban100~\cite{huang2015single}} &  \multicolumn{2}{c|}{Manga109~\cite{Manga109}}
\\
\cline{4-14}
&  &  & PSNR & SSIM & PSNR & SSIM & PSNR & SSIM & PSNR & SSIM & PSNR & SSIM 
\\
\hline
\hline
CARN~\cite{ahn2018CARN} & 1,592K & 222.8G
& 37.76
& 0.9590
& 33.52
& 0.9166
& 32.09
& 0.8978
& 31.92
& 0.9256
& 38.36
& 0.9765
\\
FALSR-A~\cite{chu2021fast} & 1,021K & 234.7G
& 37.82
& 0.959
& 33.55
& 0.9168
& 32.1
& 0.8987 
& 31.93 
& 0.9256
& -
& -
\\
IMDN~\cite{hui2019imdn} & 694K & 158.8G
& 38.00
& 0.9605
& 33.63
& 0.9177
& 32.19
& 0.8996
& 32.17
& 0.9283
& 38.88
& 0.9774
\\
LAPAR-A~\cite{li2021lapar} & 548K & 171.0G
& 38.01
& 0.9605
& 33.62
& 0.9183
& 32.19
& 0.8999
& 32.10
& 0.9283
& 38.67
& 0.9772
\\
LatticeNet~\cite{luo2020latticenet} & 756K & 169.5G
& \B{38.15}
& 0.9610
& 33.78
& 0.9193
& 32.25
& 0.9005
& 32.43
& 0.9302
& -
& -
\\
SwinIR~\cite{liang2021swinir}  & 878K & {195.6G} %
& 38.14
& \B{0.9611}
& \B{33.86}
& \B{0.9206}
& \B{32.31}
& \B{0.9012}
& \B{32.76}
& \B{0.9340}
& \B{39.12}
& \B{0.9783}
\\
\rowcolor{Gray}\textbf{Swin2SR-s}  & 1000K & {199.0G} %
& \R{38.17}
& \R{0.9613}
& \R{33.95} 
& \R{0.9216}
& \R{32.35}
& \R{0.9024}
& \R{32.85}
& \R{0.9349}
& \R{39.32}
& \R{0.9787}
\\
\hline 
\end{tabular}}
\end{center}
\end{table}

\subsection{Real-world Image Super-Resolution}
\label{sec:real}

\begin{figure}[!h]
    \centering
    \setlength{\tabcolsep}{2.0pt}
    \begin{tabular}{cccc}
    \includegraphics[width=0.47\linewidth]{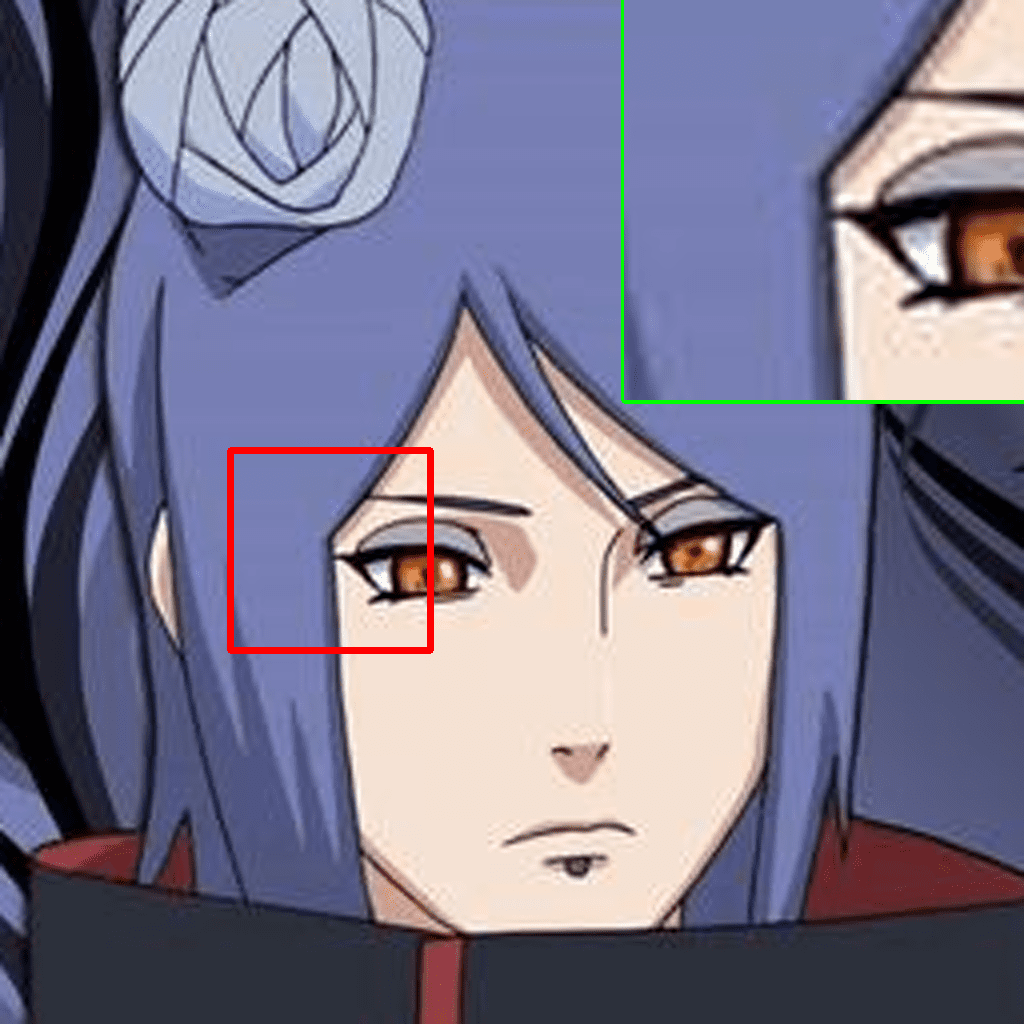} &
    \includegraphics[width=0.47\linewidth]{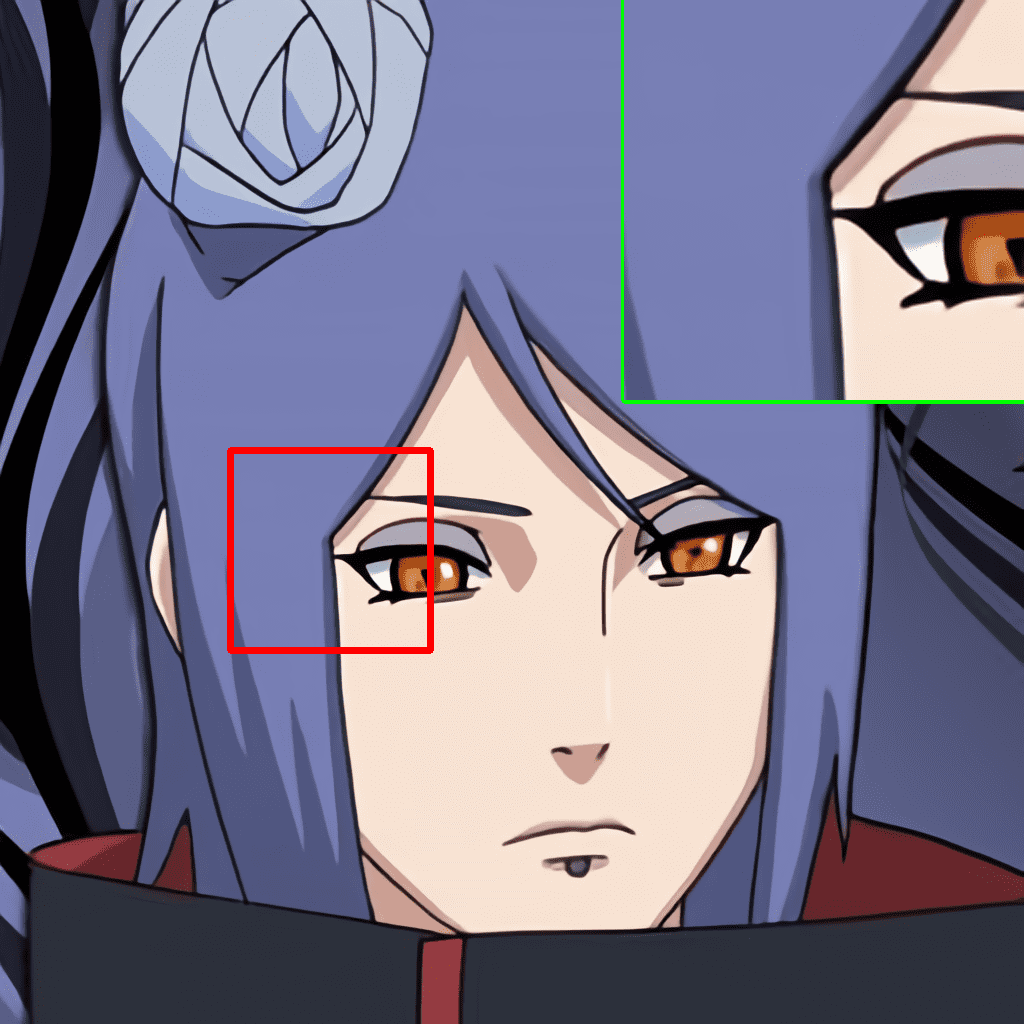}
    \tabularnewline
    \includegraphics[width=0.47\linewidth]{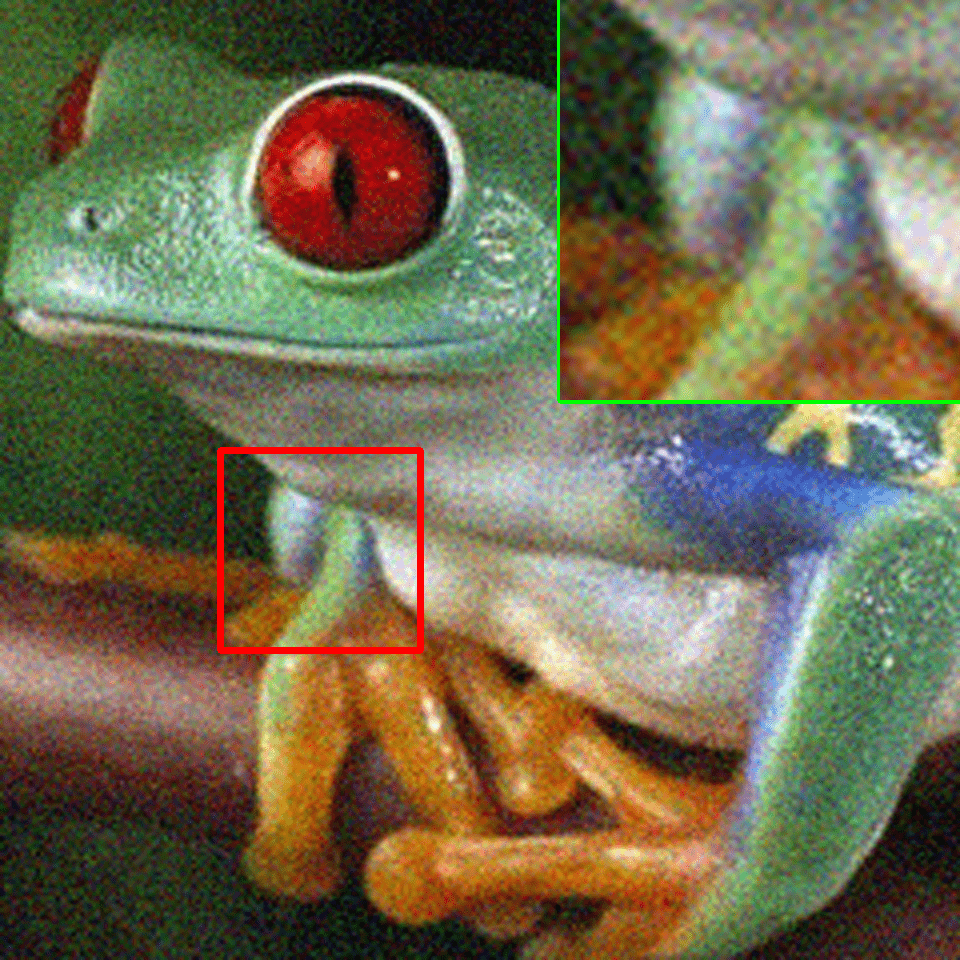} &
    \includegraphics[width=0.47\linewidth]{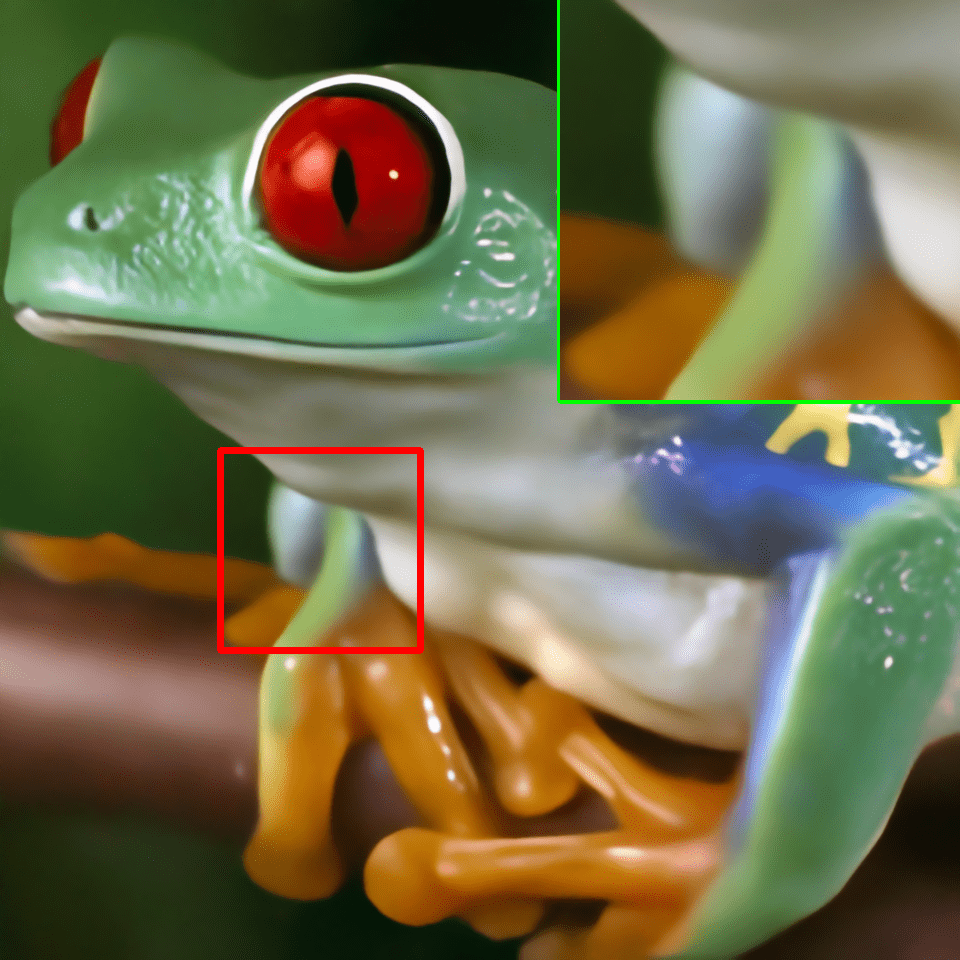}
    \tabularnewline
    \includegraphics[width=0.47\linewidth]{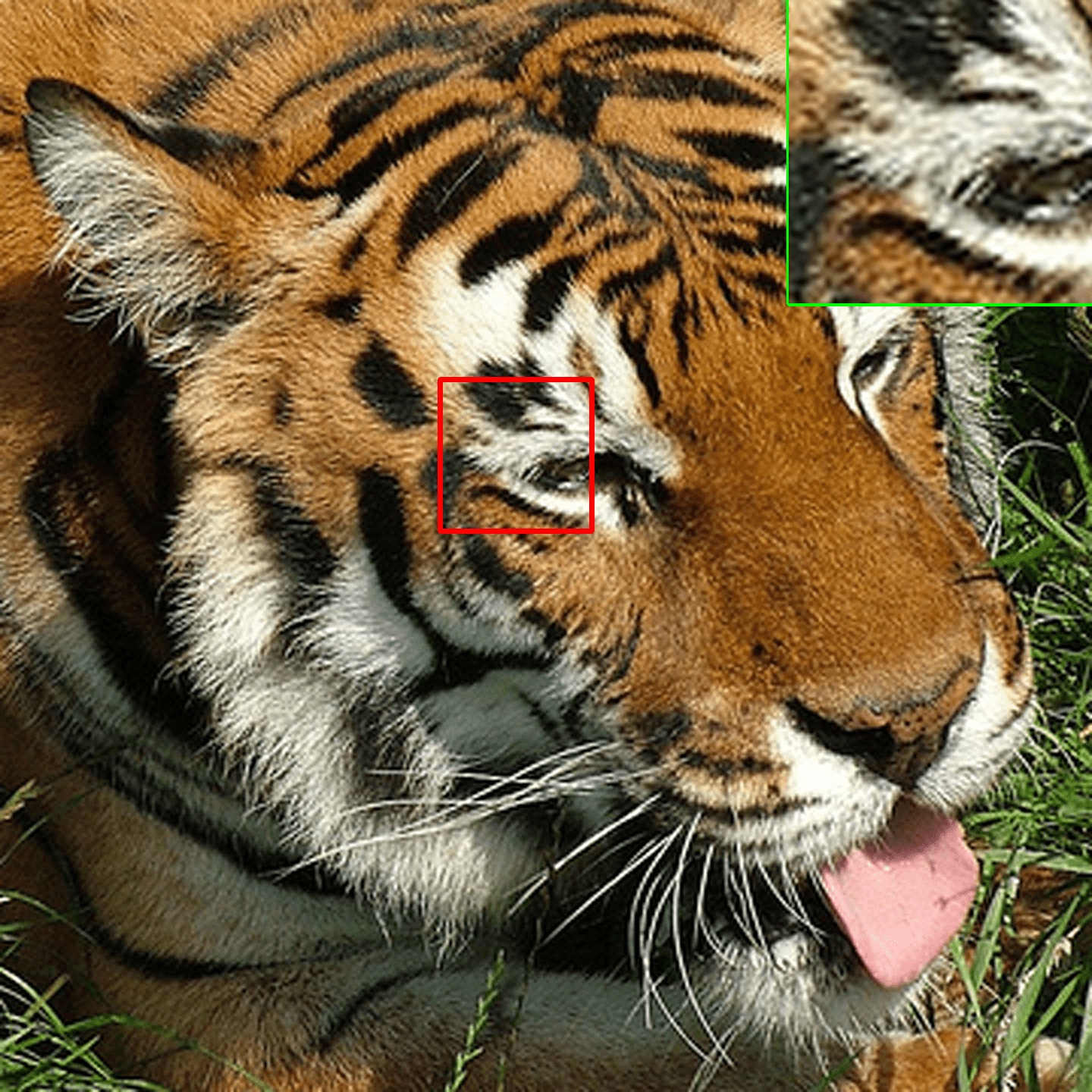} &
    \includegraphics[width=0.47\linewidth]{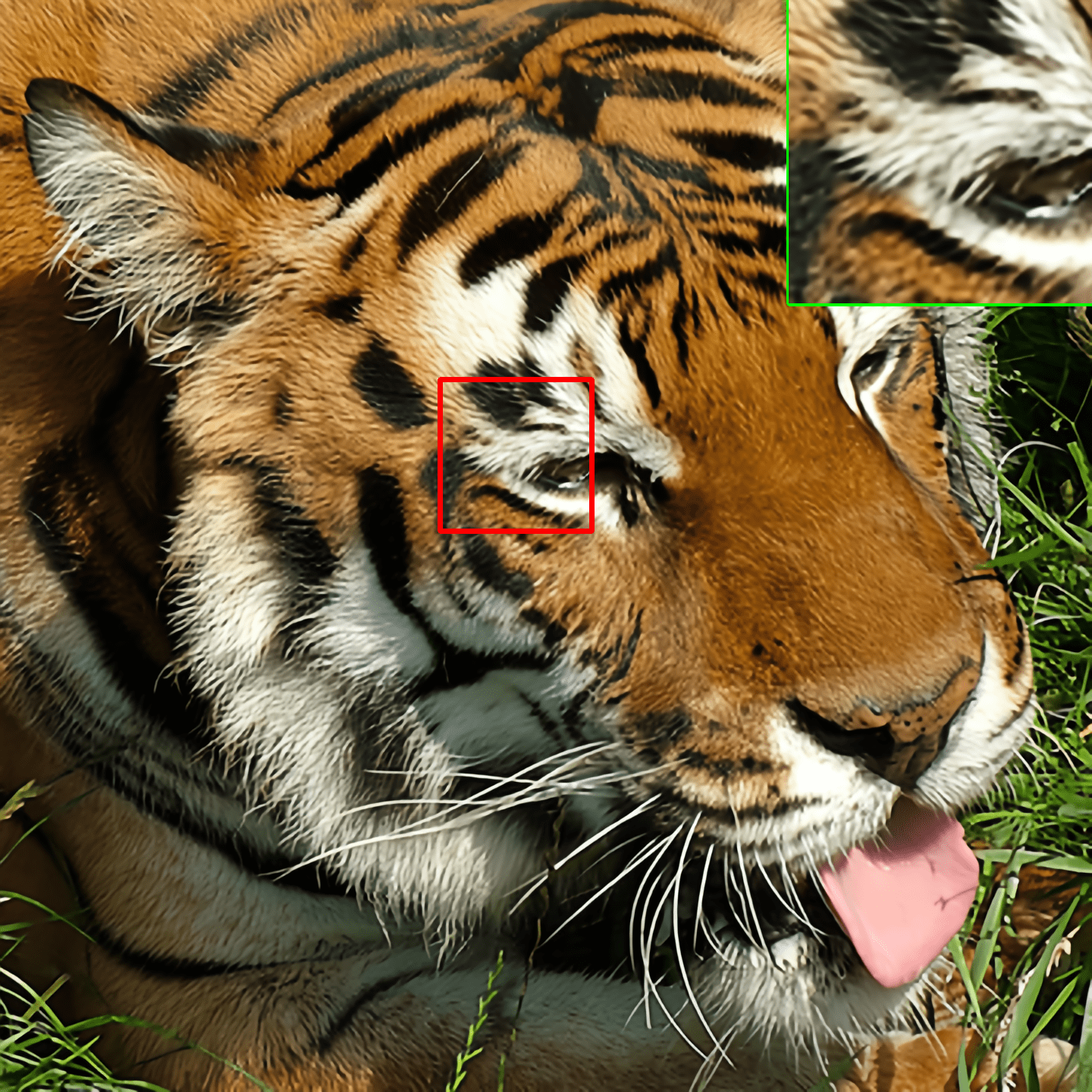}
    \tabularnewline
    LR Bicubic & Swin2SR (ours)
    \tabularnewline
    \end{tabular}
    \caption{Qualitative results on \textbf{real-world} SR datasets (RealSRSet, 5images). Our model can recover textures, remove noise and produce pleasant results.}
    \label{fig:real-sr}
\end{figure}

We also test our approach using real-world images and prove the generalization capabilities of \ours. We use the same setup as SwinIR~\cite{liang2021swinir} for training and testing our methods to exploit the full potential of these transformer-based approaches.
Since there is no ground-truth high-quality images, we only provide visual comparison with representative bicubic model in Figure~\ref{fig:real-sr}. Our model produces detailed images without artifacts. Due to the limitations of space and visualization in this document, we include the comparison with ESRGAN~\cite{wang2018esrgan} and state-of-the-art real-world image SR models such as RealSR~\cite{ji2020realsr}, BSRGAN~\cite{kai2021bsrgan}, Real-ESRGAN~\cite{wang2021realESRGAN} and SwinIR~\cite{liang2021swinir} in the supplementary material.

\subsection{Compressed Image Super-Resolution}
\label{sec:aim-challenge}

The ``AIM 2022 Challenge on Super-Resolution of Compressed Image"~\cite{yang2022aim} is a step forward for establishing a benchmark of the super-resolution of JPEG images.
In this challenge, we use the popular dataset DIV2K~\cite{agustsson2017ntire} as the training, validation and test sets.
JPEG is the most commonly used image compression standard. We target the ×4 super-resolution of the images compressed with JPEG with the quality factor of 10. Figure~\ref{fig:main} illustrates this process.
We propose two solutions for this problem based on previous Sections~\ref{sec:jpeg} and~\ref{sec:classical}:

\begin{enumerate}
    \item \textbf{Swin2SR-CI} An end-to-end model for JPEG artifacts removal and super-resolution (\ie{ Figure~\ref{fig:main}}).
    
    
    \item A 2-stage approach where first we remove JPEG compression artifacts in the LR input image using \textbf{Swin2SR-DJPEG}, and second, we upscale using \textbf{Swin2SRx4} (\ie{the model trained for Classical SR, Section~\ref{sec:classical}}). We refer to this experiment as ``Swin2SR-CI2".
    
\end{enumerate}

As we show in Table~\ref{tab:aim-results}~(\footnote{online leaderboard \url{https://codalab.lisn.upsaclay.fr/competitions/5076}}), our method is a top solution at the challenge. We trained \ours using only DIV2K~\cite{agustsson2017ntire} and Flickr2K~\cite{Flickr2K} datasets, in comparison with other teams like CASIA LCVG, which trained using 1 million images.
Our average testing time of Swin2SR model is 1.41s using single GPU A100.

In Figure~\ref{fig:aim-results} we show extensive qualitative results of compressed input super-resolution~\cite{yang2022aim}. Our model can recover information from the low-quality low-resolution input image, and generates high-resolution high-quality images. Among the limitations of our model, we can appreciate a clear blur effect, nevertheless, we find SwinIR~\cite{liang2021swinir} (and other \textit{state-of-the-art} methods) to have the same issues.

\begin{table}[!hb]
\setlength\tabcolsep{4pt}
  \centering
  \caption{Results of AIM 2022 Challenge on Super-Resolution of Compressed Image. Our solutions are placed among the top teams, while our methods can process a single image in under a second (w/o self-ensemble).}
  \vspace{2.5mm}
  \label{tab:aim-results}
  \begin{tabular}{lcccc}
    \toprule
    Team & Test PSNR (dB) & Runtime (s) & Hardware  \\
    \midrule
    VUE                        & 23.6677 & 120 & Tesla V100 \\
    BSR                        & 23.5731 & 63.96 & Tesla A100\\
    CASIA LCVG                 & 23.5597 & 78.09 &Tesla A100 \\
    USTC-IR                    & 23.5085 & 19.2 &  2080ti \\
    \rowcolor{Gray}Swin2SR-CI2 & 23.4946 & 24  & Tesla A100 \\
    MSDRSR                     & 23.4545 & 7.94 &Tesla V100 \\
    Giantpandacv               & 23.4249 & 0.248 & RTX 3090 \\
    \rowcolor{Gray}Swin2SR-CI  & 23.4033 & 9.39& Tesla A100 \\
    MVideo                     & 23.3250 & 1.7 & RTX 3090 \\
    UESTC+XJU CV               & 23.2911 & 3.0 & RTX 3090 \\
    cvlab                      & 23.2828 & 6.0 & 1080 Ti \\
    \midrule
    Bicubic $\times 4$         &22.2420  & - & -\\
    \bottomrule
    \end{tabular}%
\end{table}%

\textbf{Ensembles and fusion strategies.} We use classical self-ensemble techniques where the input image is flipped and rotated several times, and the resultant images are averaged~\cite{timofte2016seven, lugmayr2020ntire}. We only use this technique in the related AIM 2022 Challenge (Section~\ref{sec:aim-challenge} and Table~\ref{tab:aim-results}), and the marginal improvement of this technique was approximately 0.02dB PSNR.

In Table~\ref{tab:ablation} we show our ablation studies using the challenge DIV2K~\cite{agustsson2017ntire} validation set. The use of additional loss functions helped the model to converge faster, however after certain number of iterations (\ie{} 250k) the model converges. As previously mentioned, among the \textbf{limitations} of our model, we can appreciate a clear blur effect in the qualitative samples in Figure~\ref{fig:aim-results}, indicating that our model is struggling to recover fine details and sharpness.
Nevertheless, we find SwinIR~\cite{liang2021swinir} (and other \textit{state-of-the-art} methods) to have the same issues to recover the high-frequency details.
However, the overall results look very impressive considering the level of degradation of the input image ($\time4$ downsampled and compressed using JPEG at quality $q=10$). We also provide additional results and samples for DIV2K~\cite{agustsson2017ntire} in the supplementary material.

\vspace{-5mm}

\begin{table}[!ht]
    \centering
    \caption{Ablation study of our experiments in the AIM 2022 Compressed Image Super-Resolution Challenge. The additional loss functions, and our new design \ours help to converge faster and produce competitive results. Note that we compare with SwinIR pre-trained model while we trained using only the challenge DIV2K~\cite{agustsson2017ntire} data.}
    \label{tab:ablation}
    \vspace{2mm}
    \begin{tabular}{c l c}
        \toprule
         Exp. & Method & PSNR \\
         \midrule
         1 & Bicubic & 22.350 \\
         2 & RDN~\cite{zhang2018residualdense} & 23.320 \\
         3 & SwinIR~\cite{liang2021swinir} & 23.546 \\
         4 & Swin2SR~(Ours) & 23.580 \\
         5 & Swin2SR + AuxLoss & 23.585 \\
         6 & Swin2SR + AuxLoss + HFLoss & 23.590 \\
         7 & Self-ensemble Exp6 & 23.616 \\
         \bottomrule
    \end{tabular}
\end{table}

\begin{figure}[!ht]
    \centering
    \setlength{\tabcolsep}{2.0pt}
    \begin{tabular}{cc}
    \includegraphics[width=0.46\linewidth]{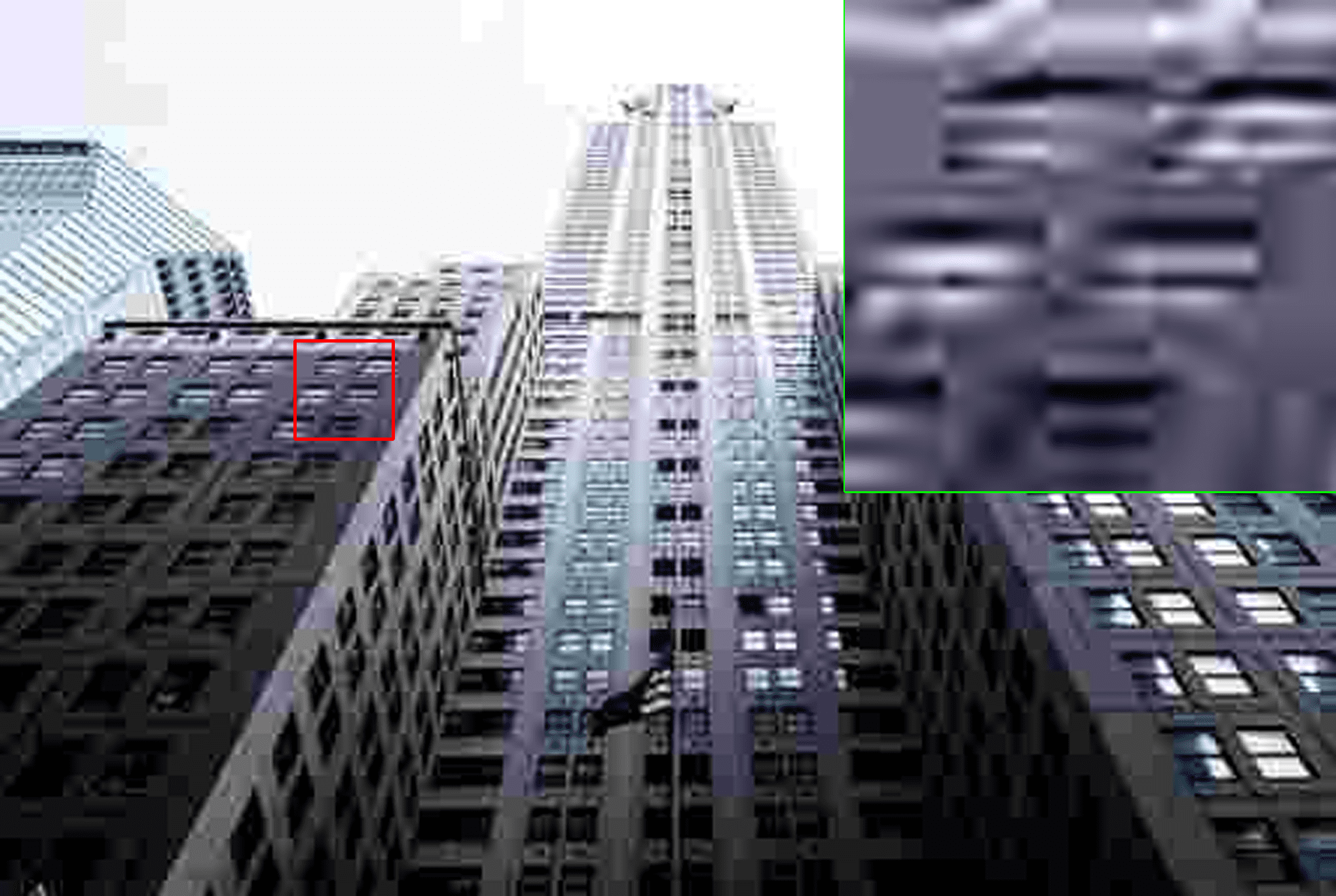} &
    \includegraphics[width=0.46\linewidth]{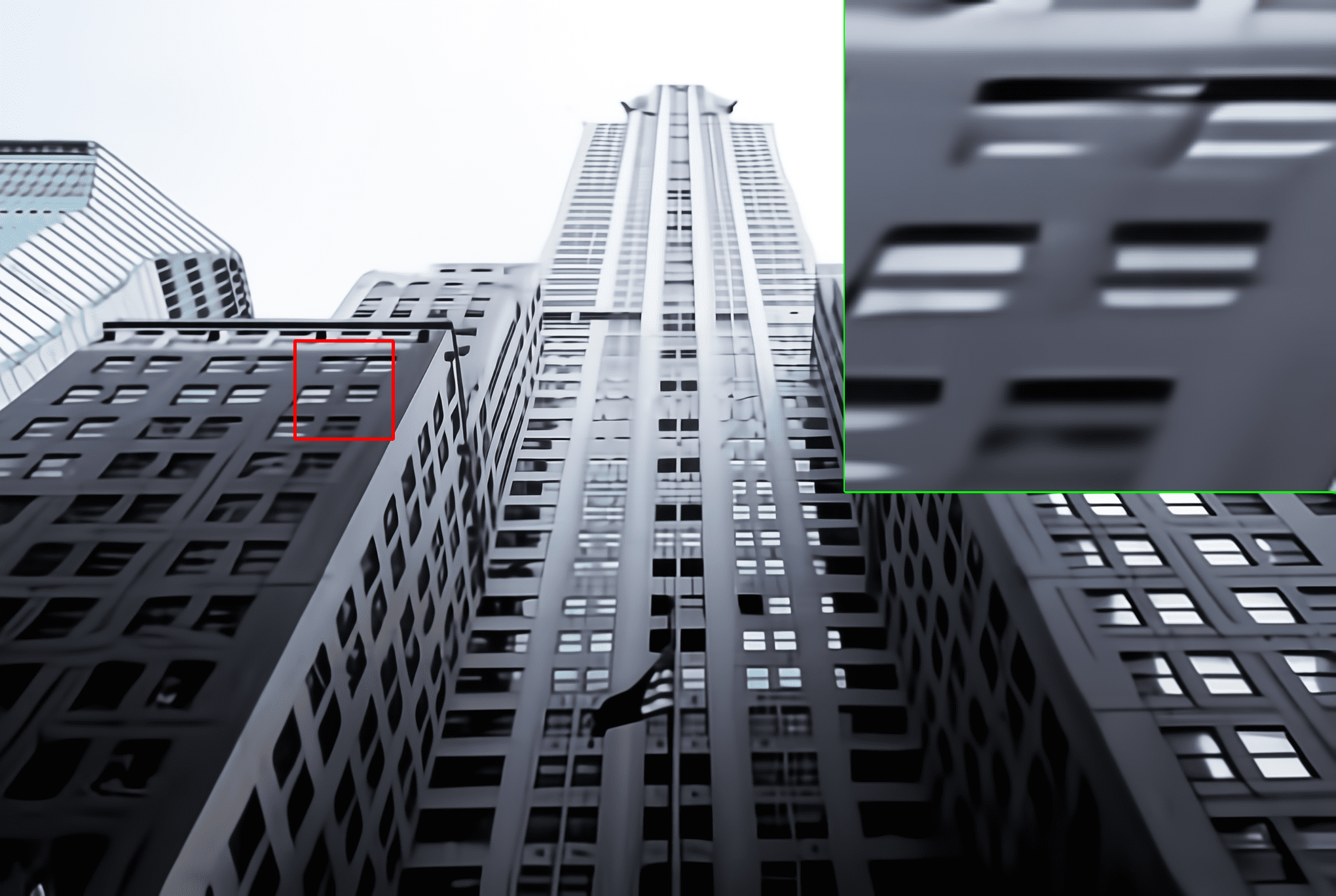}
    \tabularnewline
    Input LR ($q=40$) & SwinIR+~\cite{liang2021swinir}
    \tabularnewline
    \includegraphics[width=0.46\linewidth]{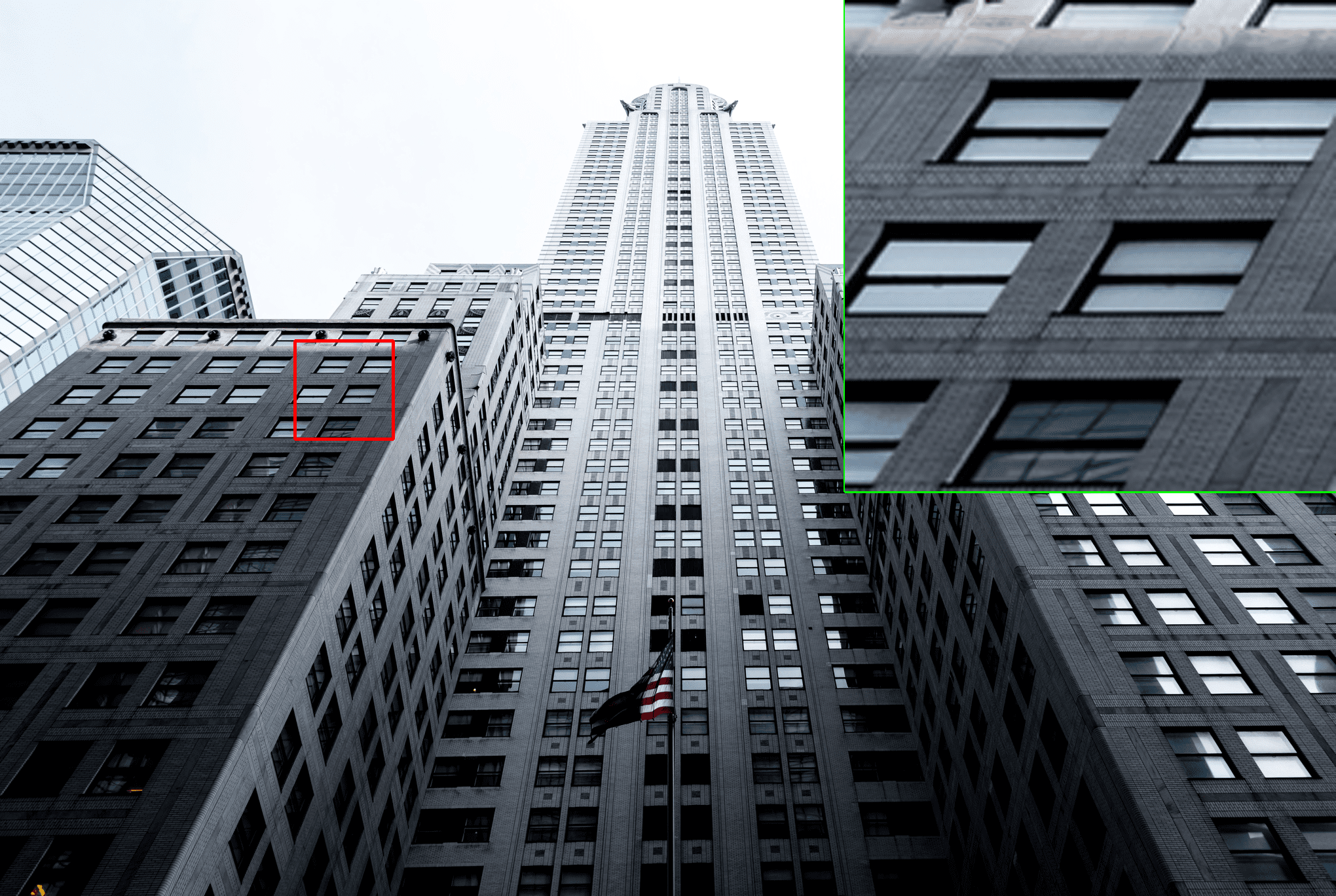} &
    \includegraphics[width=0.46\linewidth]{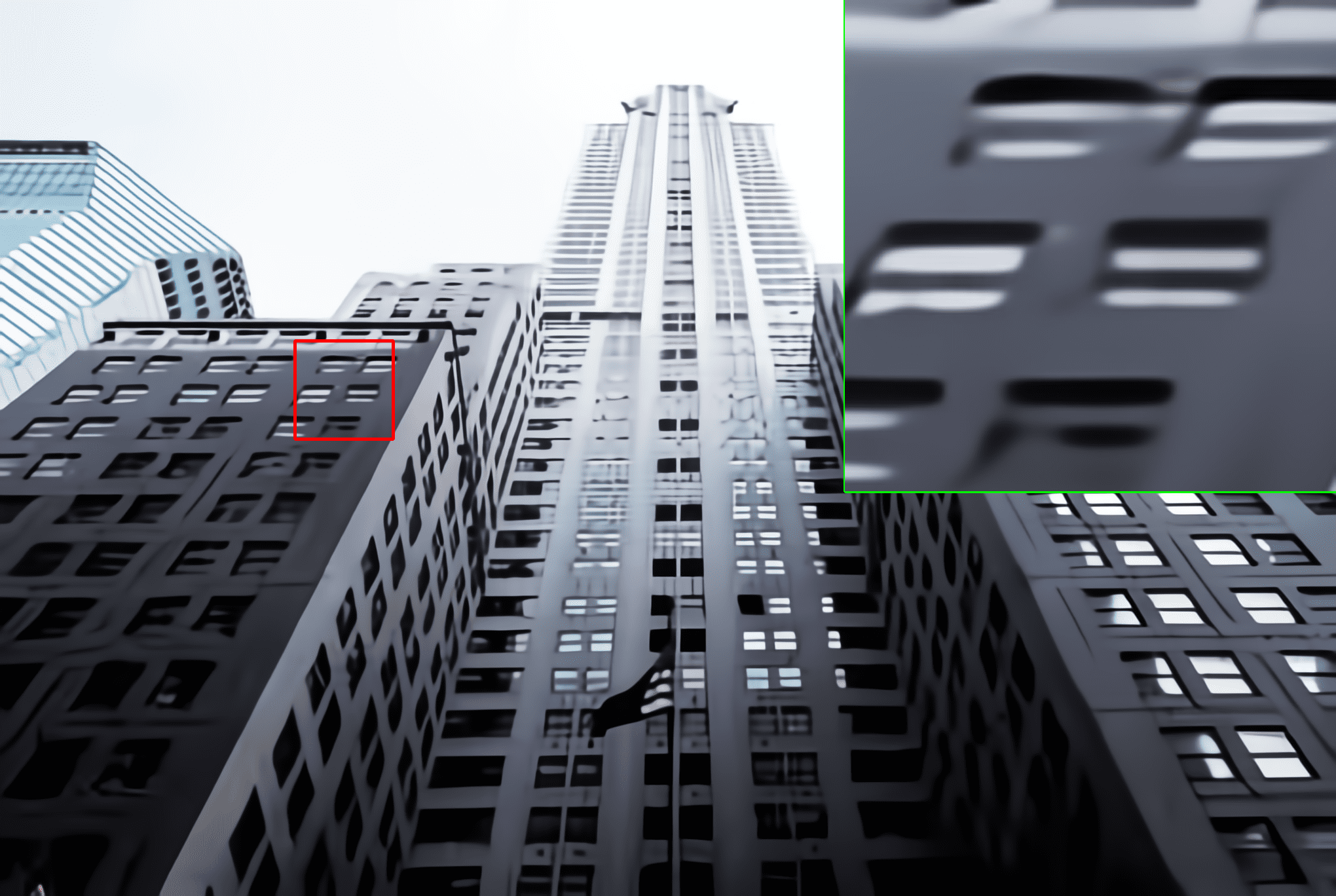}
    \tabularnewline
    Reference & Swin2SR (ours)
    \tabularnewline
    \includegraphics[width=0.46\linewidth]{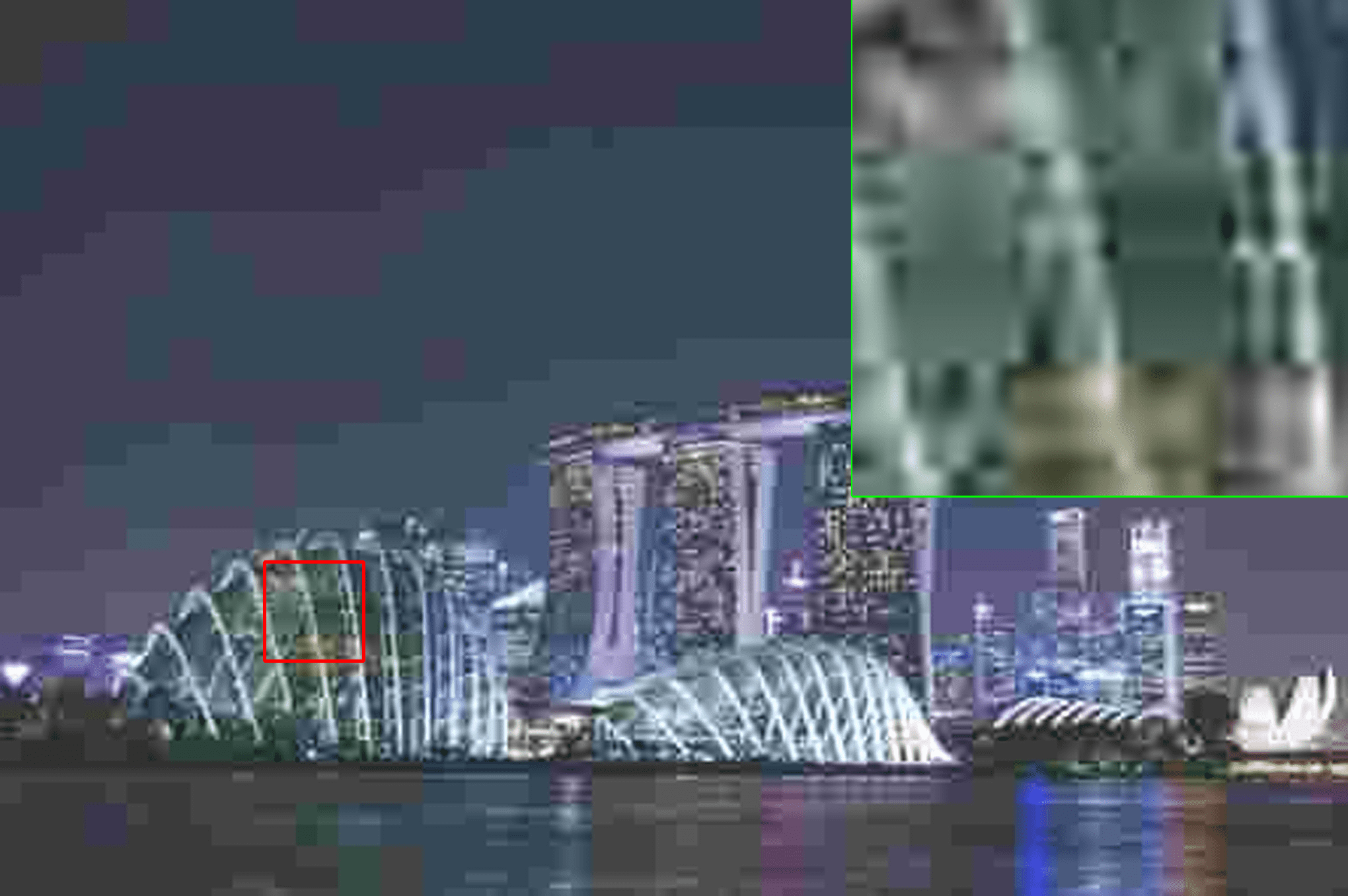} &
    \includegraphics[width=0.46\linewidth]{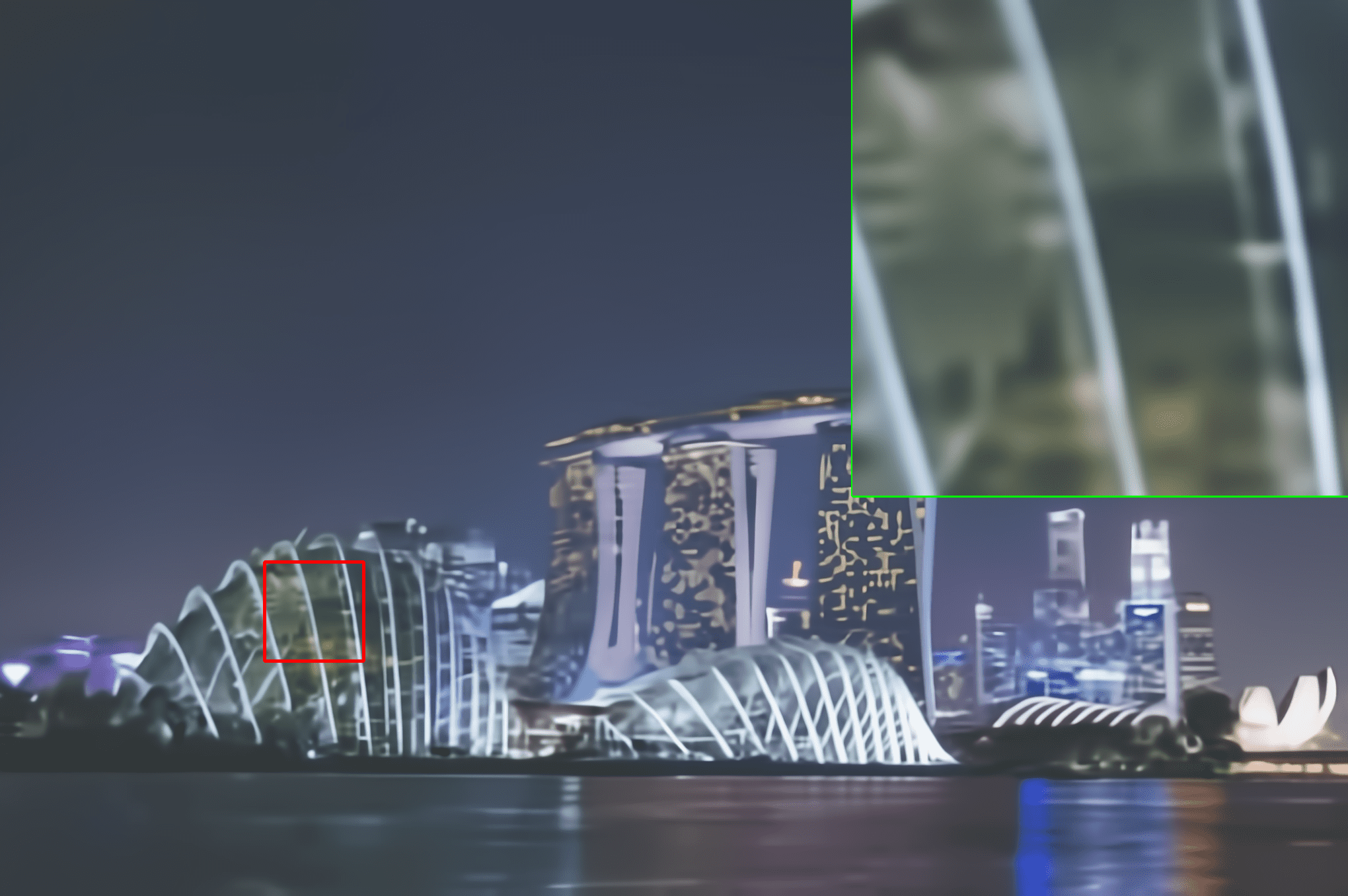}
    \tabularnewline
    Input LR ($q=40$) & SwinIR+~\cite{liang2021swinir}
    \tabularnewline
    \includegraphics[width=0.46\linewidth]{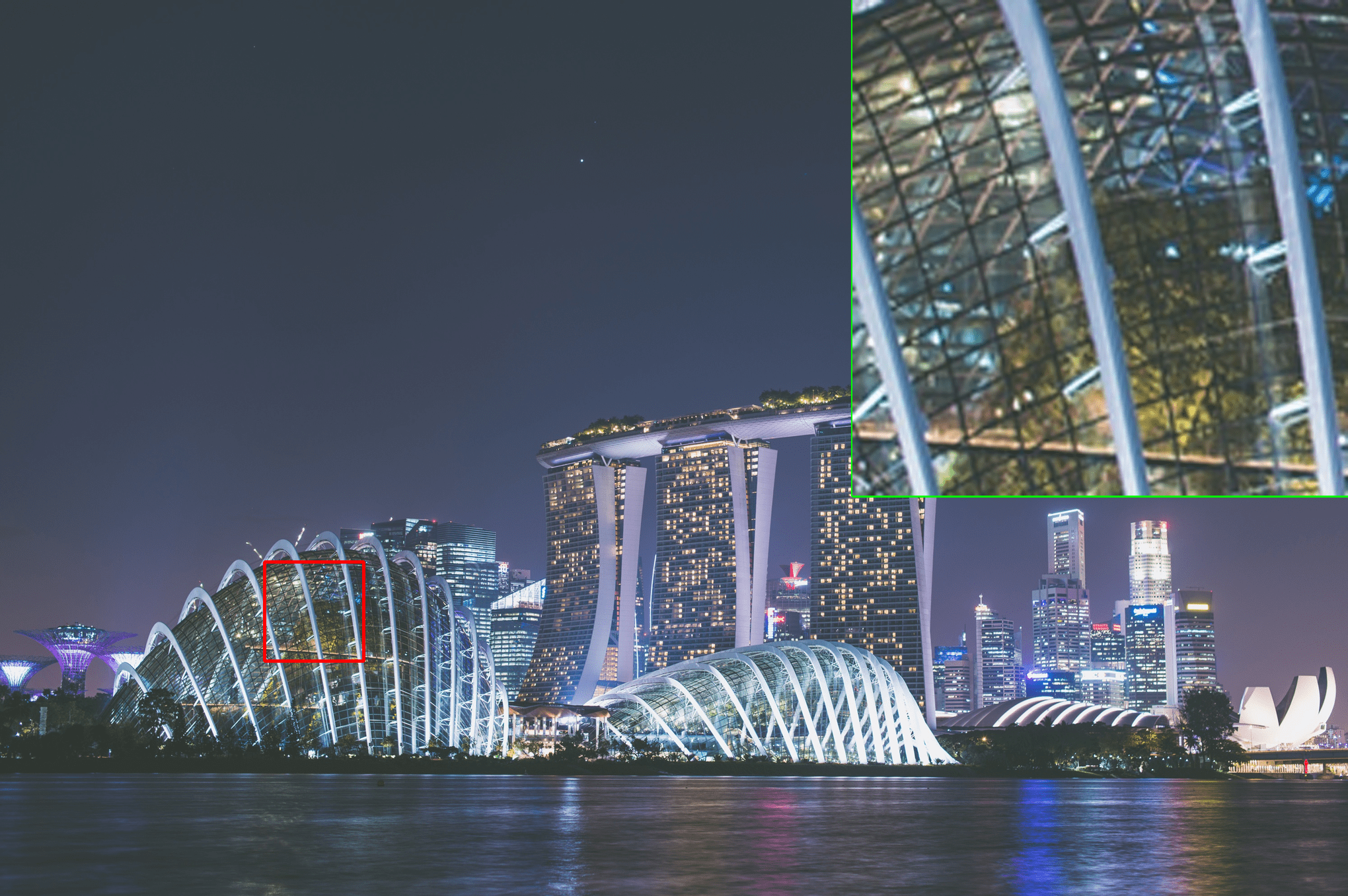} &
    \includegraphics[width=0.46\linewidth]{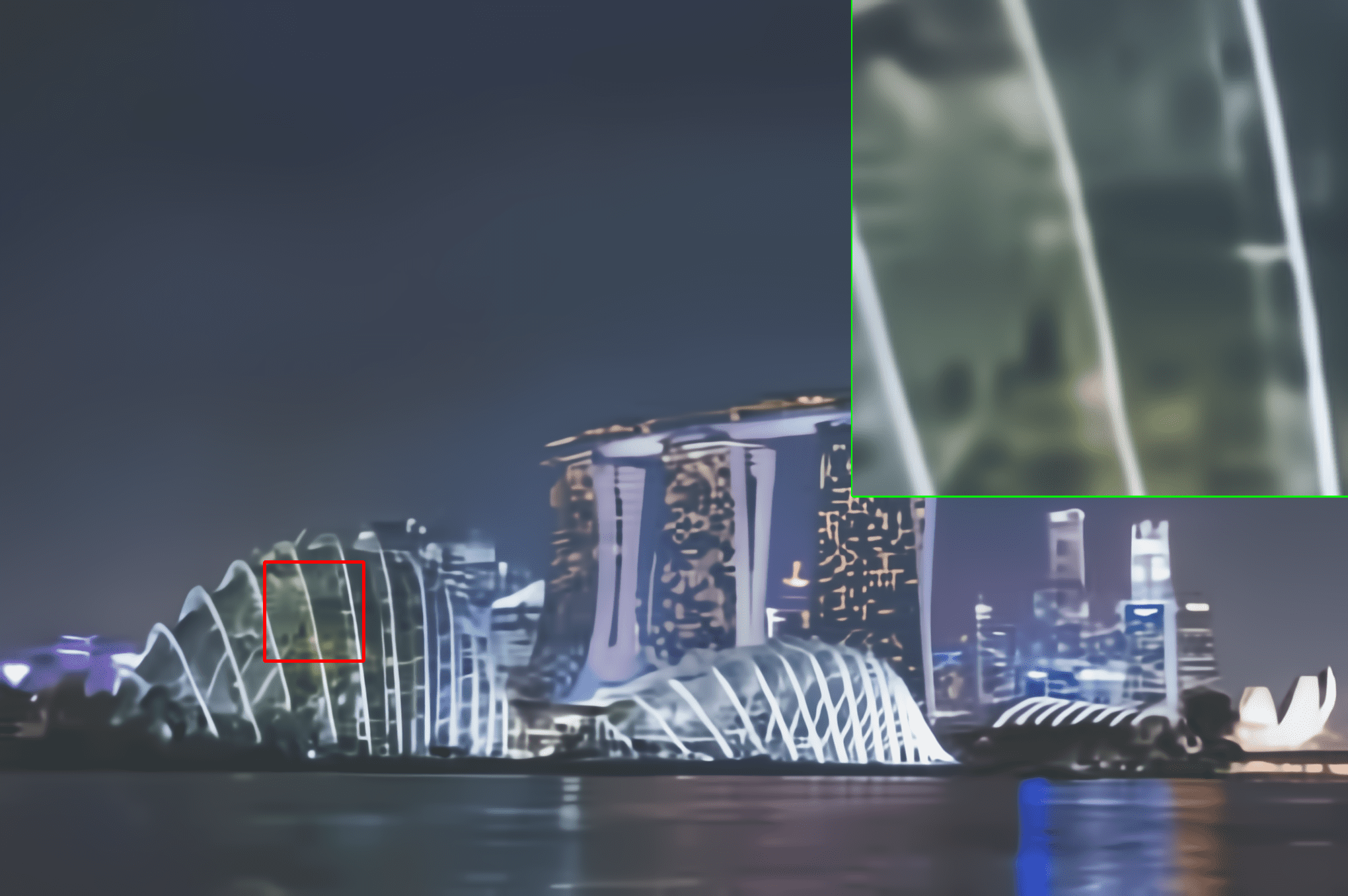}
    \tabularnewline
    Reference & Swin2SR (ours)
    \tabularnewline
    \end{tabular}
    \caption{Qualitative samples from the AIM 2022 Challenge on Super-Resolution of Compressed Image. Validation images from the DIV2K~\cite{agustsson2017ntire}.}
    \label{fig:aim-results}
\end{figure}


\section{Conclusion}
\label{sec:conclusion}

In this paper we propose Swin2SR, a SwinV2 Transformer-based model for super-resolution and restoration of compressed images. This model is a possible improvement of SwinIR (based on Swin Transformer), allowing faster training and convergence, and bigger capacity and resolution.
Extensive experiments show that Swin2SR achieves state-of-the-art performance on: JPEG compression artifacts removal, image super-resolution (classical and lightweight), and compressed image super-resolution.
Our method also achieves competitive results at the ``AIM 2022 Challenge on Super-Resolution of Compressed Image and Video", being ranked among the top-5, and therefore, it helps to advance the state-of-the-art in super-resolution of compressed inputs, which will play an essential role in industries like streaming services, virtual reality or video games.

\noindent\textbf{Acknowledgments }
This work was partly supported by The Alexander von Humboldt Foundation (AvH).

\clearpage

{\small
\bibliographystyle{plain}
\bibliography{references}

\begin{thebibliography}{10}

\bibitem{agustsson2017ntire}
Eirikur Agustsson and Radu Timofte.
\newblock Ntire 2017 challenge on single image super-resolution: Dataset and
  study.
\newblock In {\em Proceedings of the IEEE conference on computer vision and
  pattern recognition workshops}, pages 126--135, 2017.

\bibitem{ahn2018CARN}
Namhyuk Ahn, Byungkon Kang, and Kyung-Ah Sohn.
\newblock Fast, accurate, and lightweight super-resolution with cascading
  residual network.
\newblock In {\em European Conference on Computer Vision}, pages 252--268,
  2018.

\bibitem{Set5}
Marco Bevilacqua, Aline Roumy, Christine Guillemot, and Marie line
  Alberi~Morel.
\newblock Low-complexity single-image super-resolution based on nonnegative
  neighbor embedding.
\newblock In {\em British Machine Vision Conference}, pages 135.1--135.10,
  2012.

\bibitem{cao2021swinunet}
Hu~Cao, Yueyue Wang, Joy Chen, Dongsheng Jiang, Xiaopeng Zhang, Qi~Tian, and
  Manning Wang.
\newblock Swin-unet: Unet-like pure transformer for medical image segmentation.
\newblock {\em arXiv preprint arXiv:2105.05537}, 2021.

\bibitem{cao2021videosr}
Jiezhang Cao, Yawei Li, Kai Zhang, and Luc Van~Gool.
\newblock Video super-resolution transformer.
\newblock {\em arXiv preprint arXiv:2106.06847}, 2021.

\bibitem{carion2020DETR}
Nicolas Carion, Francisco Massa, Gabriel Synnaeve, Nicolas Usunier, Alexander
  Kirillov, and Sergey Zagoruyko.
\newblock End-to-end object detection with transformers.
\newblock In {\em European Conference on Computer Vision}, pages 213--229.
  Springer, 2020.

\bibitem{cavigelli2017cas}
Lukas Cavigelli, Pascal Hager, and Luca Benini.
\newblock Cas-cnn: A deep convolutional neural network for image compression
  artifact suppression.
\newblock In {\em 2017 International Joint Conference on Neural Networks},
  pages 752--759, 2017.

\bibitem{chen2021IPT}
Hanting Chen, Yunhe Wang, Tianyu Guo, Chang Xu, Yiping Deng, Zhenhua Liu, Siwei
  Ma, Chunjing Xu, Chao Xu, and Wen Gao.
\newblock Pre-trained image processing transformer.
\newblock In {\em IEEE Conference on Computer Vision and Pattern Recognition},
  pages 12299--12310, 2021.

\bibitem{chu2021fast}
Xiangxiang Chu, Bo~Zhang, Hailong Ma, Ruijun Xu, and Qingyuan Li.
\newblock Fast, accurate and lightweight super-resolution with neural
  architecture search.
\newblock In {\em International Conference on Pattern Recognition}, pages
  59--64. IEEE, 2020.

\bibitem{conde2022conformer}
Marcos~V Conde, Maxime Burchi, and Radu Timofte.
\newblock Conformer and blind noisy students for improved image quality
  assessment.
\newblock In {\em Proceedings of the IEEE/CVF Conference on Computer Vision and
  Pattern Recognition}, pages 940--950, 2022.

\bibitem{conde2022swin2sr}
Marcos~V Conde, Ui-Jin Choi, Maxime Burchi, and Radu Timofte.
\newblock {S}win2{SR}: Swinv2 transformer for compressed image super-resolution
  and restoration.
\newblock In {\em Proceedings of the European Conference on Computer Vision
  (ECCV) Workshops}, 2022.

\bibitem{conde2022modelbased}
Marcos~V. Conde, Steven McDonagh, Matteo Maggioni, Ales Leonardis, and Eduardo
  Pérez-Pellitero.
\newblock Model-based image signal processors via learnable dictionaries.
\newblock {\em Proceedings of the AAAI Conference on Artificial Intelligence},
  36(1):481--489, Jun. 2022.

\bibitem{conde2021clip}
Marcos~V Conde and Kerem Turgutlu.
\newblock Clip-art: contrastive pre-training for fine-grained art
  classification.
\newblock In {\em Proceedings of the IEEE/CVF Conference on Computer Vision and
  Pattern Recognition}, pages 3956--3960, 2021.

\bibitem{conde2021exploring}
Marcos~V Conde and Kerem Turgutlu.
\newblock Exploring vision transformers for fine-grained classification.
\newblock {\em arXiv preprint arXiv:2106.10587}, 2021.

\bibitem{dai2019SAN}
Tao Dai, Jianrui Cai, Yongbing Zhang, Shu-Tao Xia, and Lei Zhang.
\newblock Second-order attention network for single image super-resolution.
\newblock In {\em IEEE Conference on Computer Vision and Pattern Recognition},
  pages 11065--11074, 2019.

\bibitem{dong2015compression}
Chao Dong, Yubin Deng, Chen~Change Loy, and Xiaoou Tang.
\newblock Compression artifacts reduction by a deep convolutional network.
\newblock In {\em IEEE International Conference on Computer Vision}, pages
  576--584, 2015.

\bibitem{dong2014srcnn}
Chao Dong, Chen~Change Loy, Kaiming He, and Xiaoou Tang.
\newblock Learning a deep convolutional network for image super-resolution.
\newblock In {\em European Conference on Computer Vision}, pages 184--199,
  2014.

\bibitem{dosovitskiy2020ViT}
Alexey Dosovitskiy, Lucas Beyer, Alexander Kolesnikov, Dirk Weissenborn,
  Xiaohua Zhai, Thomas Unterthiner, Mostafa Dehghani, Matthias Minderer, Georg
  Heigold, Sylvain Gelly, et~al.
\newblock An image is worth 16x16 words: Transformers for image recognition at
  scale.
\newblock {\em arXiv preprint arXiv:2010.11929}, 2020.

\bibitem{ehrlich2020quantization}
Max Ehrlich, Larry Davis, Ser-Nam Lim, and Abhinav Shrivastava.
\newblock Quantization guided jpeg artifact correction.
\newblock In {\em European Conference on Computer Vision}, pages 293--309,
  2020.

\bibitem{foi2007Classic5}
Alessandro Foi, Vladimir Katkovnik, and Karen Egiazarian.
\newblock Pointwise shape-adaptive dct for high-quality denoising and
  deblocking of grayscale and color images.
\newblock {\em IEEE Transactions on Image Processing}, 16(5):1395--1411, 2007.

\bibitem{fritsche2019frequency}
Manuel Fritsche, Shuhang Gu, and Radu Timofte.
\newblock Frequency separation for real-world super-resolution.
\newblock In {\em IEEE Conference on International Conference on Computer
  Vision Workshops}, pages 3599--3608, 2019.

\bibitem{gu2022ntire}
Jinjin Gu, Haoming Cai, Chao Dong, Jimmy~S Ren, Radu Timofte, Yuan Gong,
  Shanshan Lao, Shuwei Shi, Jiahao Wang, Sidi Yang, et~al.
\newblock Ntire 2022 challenge on perceptual image quality assessment.
\newblock In {\em Proceedings of the IEEE/CVF Conference on Computer Vision and
  Pattern Recognition}, pages 951--967, 2022.

\bibitem{gu2019sftmdikc}
Jinjin Gu, Hannan Lu, Wangmeng Zuo, and Chao Dong.
\newblock Blind super-resolution with iterative kernel correction.
\newblock In {\em IEEE Conference on Computer Vision and Pattern Recognition},
  pages 1604--1613, 2019.

\bibitem{haris2018DBPN}
Muhammad Haris, Gregory Shakhnarovich, and Norimichi Ukita.
\newblock Deep back-projection networks for super-resolution.
\newblock In {\em IEEE Conference on Computer Vision and Pattern Recognition},
  pages 1664--1673, 2018.

\bibitem{huang2015single}
Jia-Bin Huang, Abhishek Singh, and Narendra Ahuja.
\newblock Single image super-resolution from transformed self-exemplars.
\newblock In {\em IEEE Conference on Computer Vision and Pattern Recognition},
  pages 5197--5206, 2015.

\bibitem{hui2019imdn}
Zheng Hui, Xinbo Gao, Yunchu Yang, and Xiumei Wang.
\newblock Lightweight image super-resolution with information
  multi-distillation network.
\newblock In {\em ACM International Conference on Multimedia}, pages
  2024--2032, 2019.

\bibitem{ji2020realsr}
Xiaozhong Ji, Yun Cao, Ying Tai, Chengjie Wang, Jilin Li, and Feiyue Huang.
\newblock Real-world super-resolution via kernel estimation and noise
  injection.
\newblock In {\em IEEE Conference on Computer Vision and Pattern Recognition
  Workshops}, pages 466--467, 2020.

\bibitem{jiang2021towards}
Jiaxi Jiang, Kai Zhang, and Radu Timofte.
\newblock Towards flexible blind jpeg artifacts removal.
\newblock In {\em Proceedings of the IEEE/CVF International Conference on
  Computer Vision}, pages 4997--5006, 2021.

\bibitem{kim2016vdsr}
Jiwon Kim, Jung Kwon~Lee, and Kyoung Mu~Lee.
\newblock Accurate image super-resolution using very deep convolutional
  networks.
\newblock In {\em IEEE Conference on Computer Vision and Pattern Recognition},
  pages 1646--1654, 2016.

\bibitem{lai2017LapSRN}
Wei-Sheng Lai, Jia-Bin Huang, Narendra Ahuja, and Ming-Hsuan Yang.
\newblock Deep laplacian pyramid networks for fast and accurate
  super-resolution.
\newblock In {\em IEEE Conference on Computer Vision and Pattern Recognition},
  pages 624--632, 2017.

\bibitem{li2021lapar}
Wenbo Li, Kun Zhou, Lu~Qi, Nianjuan Jiang, Jiangbo Lu, and Jiaya Jia.
\newblock Lapar: Linearly-assembled pixel-adaptive regression network for
  single image super-resolution and beyond.
\newblock {\em arXiv preprint arXiv:2105.10422}, 2021.

\bibitem{li2019SRFBN}
Zhen Li, Jinglei Yang, Zheng Liu, Xiaomin Yang, Gwanggil Jeon, and Wei Wu.
\newblock Feedback network for image super-resolution.
\newblock In {\em IEEE Conference on Computer Vision and Pattern Recognition},
  pages 3867--3876, 2019.

\bibitem{liang2021swinir}
Jingyun Liang, Jiezhang Cao, Guolei Sun, Kai Zhang, Luc Van~Gool, and Radu
  Timofte.
\newblock Swinir: Image restoration using swin transformer.
\newblock In {\em Proceedings of the IEEE/CVF International Conference on
  Computer Vision}, pages 1833--1844, 2021.

\bibitem{liu2018NLRN}
Ding Liu, Bihan Wen, Yuchen Fan, Chen~Change Loy, and Thomas~S Huang.
\newblock Non-local recurrent network for image restoration.
\newblock {\em arXiv preprint arXiv:1806.02919}, 2018.

\bibitem{liu2018multi}
Pengju Liu, Hongzhi Zhang, Kai Zhang, Liang Lin, and Wangmeng Zuo.
\newblock Multi-level wavelet-cnn for image restoration.
\newblock In {\em Proceedings of the IEEE conference on computer vision and
  pattern recognition workshops}, pages 773--782, 2018.

\bibitem{liu2022swin}
Ze~Liu, Han Hu, Yutong Lin, Zhuliang Yao, Zhenda Xie, Yixuan Wei, Jia Ning, Yue
  Cao, Zheng Zhang, Li~Dong, et~al.
\newblock Swin transformer v2: Scaling up capacity and resolution.
\newblock In {\em Proceedings of the IEEE/CVF Conference on Computer Vision and
  Pattern Recognition}, pages 12009--12019, 2022.

\bibitem{liu2021swin}
Ze~Liu, Yutong Lin, Yue Cao, Han Hu, Yixuan Wei, Zheng Zhang, Stephen Lin, and
  Baining Guo.
\newblock Swin transformer: Hierarchical vision transformer using shifted
  windows.
\newblock In {\em Proceedings of the IEEE/CVF International Conference on
  Computer Vision}, pages 10012--10022, 2021.

\bibitem{lugmayr2020ntire}
Andreas Lugmayr, Martin Danelljan, and Radu Timofte.
\newblock Ntire 2020 challenge on real-world image super-resolution: Methods
  and results.
\newblock In {\em Proceedings of the IEEE/CVF Conference on Computer Vision and
  Pattern Recognition Workshops}, pages 494--495, 2020.

\bibitem{luo2020latticenet}
Xiaotong Luo, Yuan Xie, Yulun Zhang, Yanyun Qu, Cuihua Li, and Yun Fu.
\newblock Latticenet: Towards lightweight image super-resolution with lattice
  block.
\newblock In {\em European Conference on Computer Vision}, pages 272--289,
  2020.

\bibitem{BSD100}
David Martin, Charless Fowlkes, Doron Tal, and Jitendra Malik.
\newblock A database of human segmented natural images and its application to
  evaluating segmentation algorithms and measuring ecological statistics.
\newblock In {\em IEEE Conference on International Conference on Computer
  Vision}, pages 416--423, 2001.

\bibitem{Manga109}
Yusuke Matsui, Kota Ito, Yuji Aramaki, Azuma Fujimoto, Toru Ogawa, Toshihiko
  Yamasaki, and Kiyoharu Aizawa.
\newblock Sketch-based manga retrieval using manga109 dataset.
\newblock {\em Multimedia Tools and Applications}, 76(20):21811--21838, 2017.

\bibitem{mei2021NLSA}
Yiqun Mei, Yuchen Fan, and Yuqian Zhou.
\newblock Image super-resolution with non-local sparse attention.
\newblock In {\em IEEE Conference on Computer Vision and Pattern Recognition},
  pages 3517--3526, 2021.

\bibitem{niu2020HAN}
Ben Niu, Weilei Wen, Wenqi Ren, Xiangde Zhang, Lianping Yang, Shuzhen Wang,
  Kaihao Zhang, Xiaochun Cao, and Haifeng Shen.
\newblock Single image super-resolution via a holistic attention network.
\newblock In {\em European Conference on Computer Vision}, pages 191--207,
  2020.

\bibitem{icb}
Rawzor.
\newblock Image compression benchmark.

\bibitem{sheikh2005live}
HR~Sheikh.
\newblock Live image quality assessment database release 2.
\newblock {\em http://live. ece. utexas. edu/research/quality}, 2005.

\bibitem{tai2017memnet}
Ying Tai, Jian Yang, Xiaoming Liu, and Chunyan Xu.
\newblock Memnet: A persistent memory network for image restoration.
\newblock In {\em IEEE International Conference on Computer Vision}, pages
  4539--4547, 2017.

\bibitem{Flickr2K}
Radu Timofte, Eirikur Agustsson, Luc Van~Gool, Ming-Hsuan Yang, and Lei Zhang.
\newblock Ntire 2017 challenge on single image super-resolution: Methods and
  results.
\newblock In {\em IEEE Conference on Computer Vision and Pattern Recognition
  Workshops}, pages 114--125, 2017.

\bibitem{timofte2013anchored}
Radu Timofte, Vincent De~Smet, and Luc Van~Gool.
\newblock Anchored neighborhood regression for fast example-based
  super-resolution.
\newblock In {\em IEEE Conference on International Conference on Computer
  Vision}, pages 1920--1927, 2013.

\bibitem{timofte2014a}
Radu Timofte, Vincent De~Smet, and Luc Van~Gool.
\newblock A+: Adjusted anchored neighborhood regression for fast
  super-resolution.
\newblock In {\em Asian Conference on Computer Vision}, pages 111--126, 2014.

\bibitem{timofte2016seven}
Radu Timofte, Rasmus Rothe, and Luc Van~Gool.
\newblock Seven ways to improve example-based single image super resolution.
\newblock In {\em Proceedings of the IEEE conference on computer vision and
  pattern recognition}, pages 1865--1873, 2016.

\bibitem{touvron2020DeiT}
Hugo Touvron, Matthieu Cord, Matthijs Douze, Francisco Massa, Alexandre
  Sablayrolles, and Herv{\'e} J{\'e}gou.
\newblock Training data-efficient image transformers \& distillation through
  attention.
\newblock {\em arXiv preprint arXiv:2012.12877}, 2020.

\bibitem{vaswani2021SAhaloing}
Ashish Vaswani, Prajit Ramachandran, Aravind Srinivas, Niki Parmar, Blake
  Hechtman, and Jonathon Shlens.
\newblock Scaling local self-attention for parameter efficient visual
  backbones.
\newblock {\em arXiv preprint arXiv:2103.12731}, 2021.

\bibitem{vaswani2017transformer}
Ashish Vaswani, Noam Shazeer, Niki Parmar, Jakob Uszkoreit, Llion Jones,
  Aidan~N Gomez, Lukasz Kaiser, and Illia Polosukhin.
\newblock Attention is all you need.
\newblock {\em arXiv preprint arXiv:1706.03762}, 2017.

\bibitem{wang2021realESRGAN}
Xintao Wang, Liangbin Xie, Chao Dong, and Ying Shan.
\newblock Real-esrgan: Training real-world blind super-resolution with pure
  synthetic data.
\newblock {\em arXiv preprint arXiv:2107.10833}, 2021.

\bibitem{wang2018esrgan}
Xintao Wang, Ke~Yu, Shixiang Wu, Jinjin Gu, Yihao Liu, Chao Dong, Yu~Qiao, and
  Chen Change~Loy.
\newblock Esrgan: Enhanced super-resolution generative adversarial networks.
\newblock In {\em European Conference on Computer Vision Workshops}, pages
  701--710, 2018.

\bibitem{wang2021uformer}
Zhendong Wang, Xiaodong Cun, Jianmin Bao, and Jianzhuang Liu.
\newblock Uformer: A general u-shaped transformer for image restoration.
\newblock {\em arXiv preprint arXiv:2106.03106}, 2021.

\bibitem{yamac2021kernelnet}
Mehmet Yamac, Baran Ataman, and Aakif Nawaz.
\newblock Kernelnet: A blind super-resolution kernel estimation network.
\newblock In {\em Proceedings of the IEEE/CVF Conference on Computer Vision and
  Pattern Recognition Workshops (CVPRW)}, pages 453--462, 2021.

\bibitem{yang2021ntire}
Ren Yang, Radu Timofte, et~al.
\newblock {NTIRE 2021} challenge on quality enhancement of compressed video:
  Methods and results.
\newblock In {\em IEEE/CVF Conference on Computer Vision and Pattern
  Recognition (CVPR) Workshops}, 2021.

\bibitem{yang2022aim}
Ren Yang, Radu Timofte, et~al.
\newblock Aim 2022 challenge on super-resolution of compressed image and video:
  Dataset, methods and results.
\newblock In {\em Proceedings of the European Conference on Computer Vision
  Workshops (ECCVW)}, 2022.

\bibitem{yang2022ntire}
Ren Yang, Radu Timofte, et~al.
\newblock {NTIRE} 2022 challenge on super-resolution and quality enhancement of
  compressed video: Dataset, methods and results.
\newblock In {\em Proceedings of the IEEE/CVF Conference on Computer Vision and
  Pattern Recognition (CVPR) Workshops}, 2022.

\bibitem{Set14}
Roman Zeyde, Michael Elad, and Matan Protter.
\newblock On single image scale-up using sparse-representations.
\newblock In {\em International Conference on Curves and Surfaces}, pages
  711--730, 2010.

\bibitem{zhang2021DPIR}
Kai Zhang, Yawei Li, Wangmeng Zuo, Lei Zhang, Luc Van~Gool, and Radu Timofte.
\newblock Plug-and-play image restoration with deep denoiser prior.
\newblock {\em IEEE Transactions on Pattern Analysis and Machine Intelligence},
  2021.

\bibitem{kai2021bsrgan}
Kai Zhang, Jingyun Liang, Luc Van~Gool, and Radu Timofte.
\newblock Designing a practical degradation model for deep blind image
  super-resolution.
\newblock In {\em IEEE Conference on International Conference on Computer
  Vision}, 2021.

\bibitem{zhang2017DnCNN}
Kai Zhang, Wangmeng Zuo, Yunjin Chen, Deyu Meng, and Lei Zhang.
\newblock Beyond a gaussian denoiser: Residual learning of deep cnn for image
  denoising.
\newblock {\em IEEE transactions on image processing}, 26(7):3142--3155, 2017.

\bibitem{zhang2017IRCNN}
Kai Zhang, Wangmeng Zuo, Shuhang Gu, and Lei Zhang.
\newblock Learning deep cnn denoiser prior for image restoration.
\newblock In {\em IEEE Conference on Computer Vision and Pattern Recognition},
  pages 3929--3938, 2017.

\bibitem{zhang2018ffdnet}
Kai Zhang, Wangmeng Zuo, and Lei Zhang.
\newblock Ffdnet: Toward a fast and flexible solution for cnn-based image
  denoising.
\newblock {\em IEEE Transactions on Image Processing}, 27(9):4608--4622, 2018.

\bibitem{zhang2018srmd}
Kai Zhang, Wangmeng Zuo, and Lei Zhang.
\newblock Learning a single convolutional super-resolution network for multiple
  degradations.
\newblock In {\em IEEE Conference on Computer Vision and Pattern Recognition},
  pages 3262--3271, 2018.

\bibitem{zhang2018rcan}
Yulun Zhang, Kunpeng Li, Kai Li, Lichen Wang, Bineng Zhong, and Yun Fu.
\newblock Image super-resolution using very deep residual channel attention
  networks.
\newblock In {\em European Conference on Computer Vision}, pages 286--301,
  2018.

\bibitem{zhang2019RNAN}
Yulun Zhang, Kunpeng Li, Kai Li, Bineng Zhong, and Yun Fu.
\newblock Residual non-local attention networks for image restoration.
\newblock {\em arXiv preprint arXiv:1903.10082}, 2019.

\bibitem{zhang2018residualdense}
Yulun Zhang, Yapeng Tian, Yu~Kong, Bineng Zhong, and Yun Fu.
\newblock Residual dense network for image super-resolution.
\newblock In {\em Proceedings of the IEEE conference on computer vision and
  pattern recognition}, pages 2472--2481, 2018.

\bibitem{zhang2020RDNIR}
Yulun Zhang, Yapeng Tian, Yu~Kong, Bineng Zhong, and Yun Fu.
\newblock Residual dense network for image restoration.
\newblock {\em IEEE Transactions on Pattern Analysis and Machine Intelligence},
  43(7):2480--2495, 2020.

\bibitem{zheng2019implicit}
Bolun Zheng, Yaowu Chen, Xiang Tian, Fan Zhou, and Xuesong Liu.
\newblock Implicit dual-domain convolutional network for robust color image
  compression artifact reduction.
\newblock {\em IEEE Transactions on Circuits and Systems for Video Technology},
  30(11):3982--3994, 2019.

\bibitem{zhou2020IGNN}
Shangchen Zhou, Jiawei Zhang, Wangmeng Zuo, and Chen~Change Loy.
\newblock Cross-scale internal graph neural network for image super-resolution.
\newblock {\em arXiv preprint arXiv:2006.16673}, 2020.

\end{thebibliography}
}

\end{document}